\documentclass[lettersize,journal]{IEEEtran}
\usepackage{amsmath,amsfonts,amssymb}
\usepackage{algorithmic}
\usepackage{algorithm}
\usepackage{array}
\usepackage{textcomp}
\usepackage{stfloats}
\usepackage{url}
\usepackage{verbatim}
\usepackage{graphicx}
\usepackage{cite}
\usepackage{subfigure}
\usepackage{booktabs}
\usepackage{bbding}
\usepackage{makecell}
\usepackage{CJKutf8}
\usepackage{etoolbox}
\makeatletter
\patchcmd{\@makecaption}
  {\scshape}
  {}
  {}
  {}
\makeatother
\usepackage{color}
\usepackage{bm}

\newtheorem{definition}{\textbf{Definition}}[section]
\newtheorem{theorem}{\textbf{Theorem}}

\newtheorem{lemma}{\textbf{Lemma}}
\newtheorem{remark}{\textbf{Remark}}[section]
\newtheorem{assumption}{\textbf{Assumption}}

\def\A{\bm{\mathcal{A}}}
\def\B{\bm{\mathcal{B}}}
\def\C{\bm{\mathcal{C}}}
\def\D{\bm{\mathcal{D}}}
\def\E{\bm{\mathcal{E}}}
\def\F{\bm{\mathcal{F}}}
\def\G{\bm{\mathcal{G}}}
\def\H{\bm{\mathcal{H}}}
\def\I{\bm{\mathcal{I}}}

\def\K{\bm{\mathcal{K}}}
\def\L{\bm{\mathcal{L}}}
\def\M{\bm{\mathfrak{M}}}
\def\O{\bm{\mathcal{O}}}
\def\Q{\bm{\mathcal{Q}}}
\def\R{\bm{\mathcal{R}}}
\def\S{\bm{\mathcal{S}}}
\def\T{\bm{\mathcal{T}}}
\def\U{\bm{\mathcal{U}}}
\def\V{\bm{\mathcal{V}}}
\def\X{\bm{\mathcal{X}}}
\def\Y{\bm{\mathcal{Y}}}
\def\Z{\bm{\mathcal{Z}}}

\def\P{\bm{\mathcal{P}}}
\def\W{\bm{\mathcal{W}}}

\def\r{\bm{r}}
\def\y{\bm{y}}

\begin{document}

\title{Efficient Low-Tubal-Rank Tensor Estimation via \\ Alternating Preconditioned Gradient Descent}

\author{Zhiyu Liu, Zhi Han, Yandong Tang, Jun Fan, Yao Wang
\thanks{This work was supported in part by the National Natural Science Foundation of China under Grant U23A20343, 62303447, 61821005; in part by the CAS Project for Young Scientists in Basic Research under Grant YSBR-041; in part by the Liaoning Provincial ``Selecting the Best Candidates by Opening Competition Mechanism" Science and Technology Program under Grant 2023JH1/10400045;  in part by the Youth Innovation Promotion Association of the Chinese Academy of Sciences under Grant 2022196.}
\thanks{Zhiyu Liu is with the State Key
Laboratory of Robotics and Intelligent Systems, Shenyang Institute of Automation, Chinese Academy
of Sciences, Shenyang 110016, P.R. China, and also with the University of Chinese Academy
of Sciences, Beijing 100049, China (email: liuzhiyu@sia.cn).}% <-this % stops a space
\thanks{Zhi Han, Yandong Tang are with the State Key
Laboratory of Robotics and Intelligent Systems, Shenyang Institute of Automation, Chinese Academy
of Sciences, Shenyang 110016, P.R. China (email: hanzhi@sia.cn; ytang@sia.cn).}
% \thanks{Xi-Le Zhao is with the University of Electronic Science and Technology of
% China, Chengdu 610051, China (e-mail: xlzhao122003@163.com).}
\thanks{Jun Fan is with the Department of Mathematics, Hong Kong Baptist University, Hong Kong, 999077 (e-mail: junfan@hkbu.edu.hk).   }
\thanks{Yao Wang is with the Center for Intelligent Decision-making and Machine
Learning, School of Management, Xi’an Jiaotong University, Xi’an 710049,
P.R. China. (email: yao.s.wang@gmail.com).}
}

%The paper headers
% \markboth{Journal of \LaTeX\ Class Files,~Vol.~14, No.~8, August~2021}%
% {Shell \MakeLowercase{\textit{et al.}}: A Sample Article Using IEEEtran.cls for IEEE Journals}

% \IEEEpubid{0000--0000/00\$00.00~\copyright~2021 IEEE}
% Remember, if you use this you must call \IEEEpubidadjcol in the second
% column for its text to clear the IEEEpubid mark.

\maketitle

\begin{abstract}
The problem of low-tubal-rank tensor estimation is a fundamental task with wide applications across high-dimensional signal processing, machine learning, and image science. Traditional approaches tackle such a problem by performing tensor singular value decomposition, which is computationally expensive and becomes infeasible for large-scale tensors. Recent approaches address this issue by factorizing the tensor into two smaller factor tensors and solving the resulting problem using gradient descent. However, this kind of approach requires an accurate estimate of the tensor rank, and when the rank is overestimated, the convergence of gradient descent and its variants slows down significantly or even diverges. To address this problem, we propose an Alternating Preconditioned Gradient Descent (APGD) algorithm, which accelerates convergence in the over-parameterized setting by adding a preconditioning term to the original gradient and updating these two factors alternately. Based on certain geometric assumptions on the objective function, we establish linear convergence guarantees for more general low-tubal-rank tensor estimation problems. Then we further analyze the specific cases of low-tubal-rank tensor factorization and low-tubal-rank tensor recovery. Our theoretical results show that APGD achieves linear convergence even under over-parameterization, and the convergence rate is independent of the tensor condition number. Extensive simulations on synthetic data are carried out to validate our theoretical assertions.

\end{abstract}

\begin{IEEEkeywords}
low-tubal-rank tensor estimation, t-SVD, tensor factorization, over-parameterization, preconditioned.
\end{IEEEkeywords}

\section{Introduction}

Tensors, as high-dimensional generalizations of matrices, provide a natural and powerful framework for modeling multiway data that arise in numerous applications such as signal processing\cite{sidiropoulos2017tensor,nion2010tensor,cichocki2015tensor}, recommendation systems \cite{zhu2018fairness,frolov2017tensor}, and genomics \cite{amin2023tensor,fang2019tightly}. Unlike matrices that capture pairwise relationships, tensors can represent higher-order interactions, making them well-suited for complex data structures. In fact, many problems can be formulated as low-rank tensor estimation tasks, such as tensor completion \cite{zhang2016exact,wang2025robust}, tensor compressed sensing \cite{hou2021robust,wang2021generalized}, and 1-bit tensor sensing \cite{cao20241,hou2025robust}. Specifically, the goal of low-rank tensor estimation is to recover a target low-rank tensor from partial observations, which can be formulated mathematically as 

\begin{equation}
\underset{\X}{\min} \operatorname{rank}(\X),\quad \operatorname{s.t.} \Psi (\X) = \T,
\end{equation}
where $\T$ denotes the observation from the linear operator $\Psi(\cdot)$, and $\operatorname{rank}(\cdot)$ denotes the tensor rank function. For example, in the tensor sensing problem, the observation $\textbf{y}\in\mathbb{R}^m$ is obtained from the linear measurements
$$\M(\X)= [\langle \A_1, \X \rangle,\langle \A_2, \X \rangle,...,\langle \A_m, \X \rangle].$$

Although tensors are widely used in many practical applications, there remain several uncertainties in tensor algebra. One major issue is that tensors admit multiple decomposition methods, each associated with a different definition of tensor rank. For example, popular approaches include CP decomposition (also known as CANDECOMP/PARAFAC)~\cite{carroll1970analysis, harshman1970foundations}, Tucker decomposition~\cite{tucker1966some}, tensor singular value decomposition (t-SVD)~\cite{kilmer2011factorization}, and other notions of tensor rank can be found in~\cite{oseledets2011tensor,zhao2016tensor,zheng2021fully}. CP decomposition expresses a tensor as a sum of rank-one outer products, and its corresponding CP rank is the minimum number of such components. Although this definition resembles matrix rank, computing the CP rank is NP-hard. Tucker decomposition unfolds the tensor along each mode into matrices and factorizes it into a core tensor multiplied by factor matrices. The Tucker rank is defined by the ranks of the unfolded matrices. While easier to compute, the Tucker rank may not fully capture inter-mode relationships.

To address these limitations, Kilmer et al. proposed the t-SVD framework \cite{kilmer2011factorization}, which is based on tensor-tensor product (t-product), and introduced the notion of tubal-rank. This approach first transforms a third-order tensor into the Fourier domain, performs matrix SVD on each frontal slice, and then applies the inverse transform to recover the factor tensors. The resulting decomposition is structurally similar to matrix SVD and satisfies an Eckart–Young-like theorem, allowing for optimal low-tubal-rank approximation. Moreover, the tubal-rank more accurately reflects the low-rank structure of the tensor.

In this work, we focus on the problem of third-order tensor estimation under the t-SVD framework, which can be formulated as:
\begin{equation}
\begin{aligned}
 \underset{\X\in\mathbb{R}^{n_1 \times n_2 \times n_3}}{\operatorname{minimize}} f(\X)\ \operatorname{subject}\ \operatorname{to}\ \operatorname{rank}_t (\X) \le r,
\end{aligned}
\end{equation}
where $f(\X)$ denotes the loss function and $\operatorname{rank}_t(\X)$ denotes the tubal-rank of $\X$ (see Section \ref{sec:notations} for detailed definition). For this objective function, a widely used approach is to minimize the tensor nuclear norm, which was proposed by Lu et al. \cite{lu2018exact,lu2019tensor}. Along this line, many methods have further refined the regularization term to solve the low-tubal-rank tensor estimation problem, such as IR-t-TNN \cite{wang2021generalized}. However, these methods require computing the t-SVD, which is extremely time-consuming. As the tensor dimensions increase, the computational cost of these methods becomes impractical.

To address this issue, several decomposition-based methods have been proposed \cite{zhou2017tensor,du2021unifying,liu2024low}, which are high-order extensions of the Burer-Monteiro factorization developed for low-rank matrix estimation. Specifically, methods based on Burer–Monteiro (BM) factorization \cite{burer2003nonlinear} decompose a tensor $\X$ of tubal-rank $r$ into two smaller tensors, $\L \in \mathbb{R}^{n_1 \times r \times n_3}$ and $\R \in \mathbb{R}^{n_2 \times r \times n_3}$. The objective function can then be written as:
\begin{equation}
\underset{\L\in\mathbb{R}^{n_1 \times r \times n_3},\ \R\in\mathbb{R}^{n_2 \times r \times n_3}}{\operatorname{minimize}} \ f(\L*\R^\top). 
\label{equ:03}
\end{equation}

To solve this optimization problem, some methods minimize the nuclear norm of the factor tensors, while others adopt alternating minimization. However, these approaches still incur high computational costs. More recently, gradient descent and its variants have been introduced for this task. Liu et al. \cite{liu2024low} and Karnik \cite{karnik2024implicit} et al. proved that Factorized Gradient Descent (FGD) can converge to the optimal solution in low-tubal-rank tensor recovery. However, when the tensor condition number (see Section 2 for a detailed definition) is large, the convergence of FGD becomes significantly slower. To address this, Feng et al. \cite{feng2025learnable} and Wu et al. \cite{wu2025guaranteed} proposed Tensor ScaledGD to accelerate convergence under ill-conditioned settings, which updates as:
\begin{equation}
\begin{aligned}
\L_{t+1} =& \L_t-\eta\nabla_{\L}f(\L_t*\R_t^\top)*(\R_t^\top*\R_t)^{{-1}}\\
\R_{t+1} =& \R_t-\eta\nabla_{\R}f(\L_t*\R_t^\top)*(\L_t^\top*\L_t)^{{-1}}.
\end{aligned}
\label{equ:04}
\end{equation}

However, ScaledGD assumes that the tubal-rank is known in advance. In practice, the true tubal-rank is often difficult to determine accurately, or doing so requires substantial computational cost \cite{shi2021robust,zheng2023novel}. A reasonable strategy in this case is to set a slightly larger tubal-rank, leading to an over-ranked or over-parameterized model. This idea has also been explored in both matrix and previous tensor studies \cite{zhang2021preconditioned,liu2024low}. However, it is worth noting that even when the tubal-rank is accurately estimated, the model can still be over-parameterized if the tensor is not full tubal-rank\footnote{We say that a tensor $\X_\star$ with tubal-rank $r_\star$ is not full tubal-rank if, in the frequency domain, there exists at least one frontal slice whose rank is smaller than $r_\star$.
}, which is an issue not considered in previous studies. Both situations cause GD and ScaledGD to fail, and in particular, ScaledGD diverges because the two terms $\L^\top * \L$ and $\R^\top * \R$ become non-invertible, as shown in Figure 1.
 In addition, most existing studies focus on specific problems, and there is a lack of analysis for the general low-tubal-rank tensor estimation problem.
These raise an important question: \textbf{can we design an algorithm that is robust to both ill-conditioning and over-parameterization, and can also be applied to a broad class of low-tubal-rank tensor estimation problems?}

To address this problem, we propose an Alternating Preconditioned Gradient Descent (APGD) algorithm for solving over-parameterized and ill-conditioned low-tubal-rank tensor estimation problems. First, we obtain an initial point $\X_0$, which is assumed to be close to the ground truth $\X_\star$. This can be achieved through specific initialization methods such as spectral initialization. We then decompose $\X_0$ into two smaller tensors, $\L$ and $\R$. Next, we add two preconditioning terms $(\R_t^\top * \R_t + \lambda_t \I)^{{-1}}$ and $(\L_t^\top * \L_t + \lambda_t \I)^{{-1}}$ to the original gradient. These terms differ from those in ScaledGD by introducing a damping factor $\lambda_t \I$ to handle the over-ranked case. However, the presence of this damping term breaks the covariance property of the iteration, so an additional rebalancing step is required, as shown in Algorithm 2. Finally, the two factor tensors are updated alternately, which makes the algorithm more robust to the choice of the hyperparameter $\lambda_t$, eliminating the need for manual tuning. The overall procedure is summarized in Algorithm 1.

\begin{figure}[h]
\centering
\subfigure[]{
\begin{minipage}[t]{0.48\linewidth}
\centering
\includegraphics[width=4.3cm,height=4.3cm]{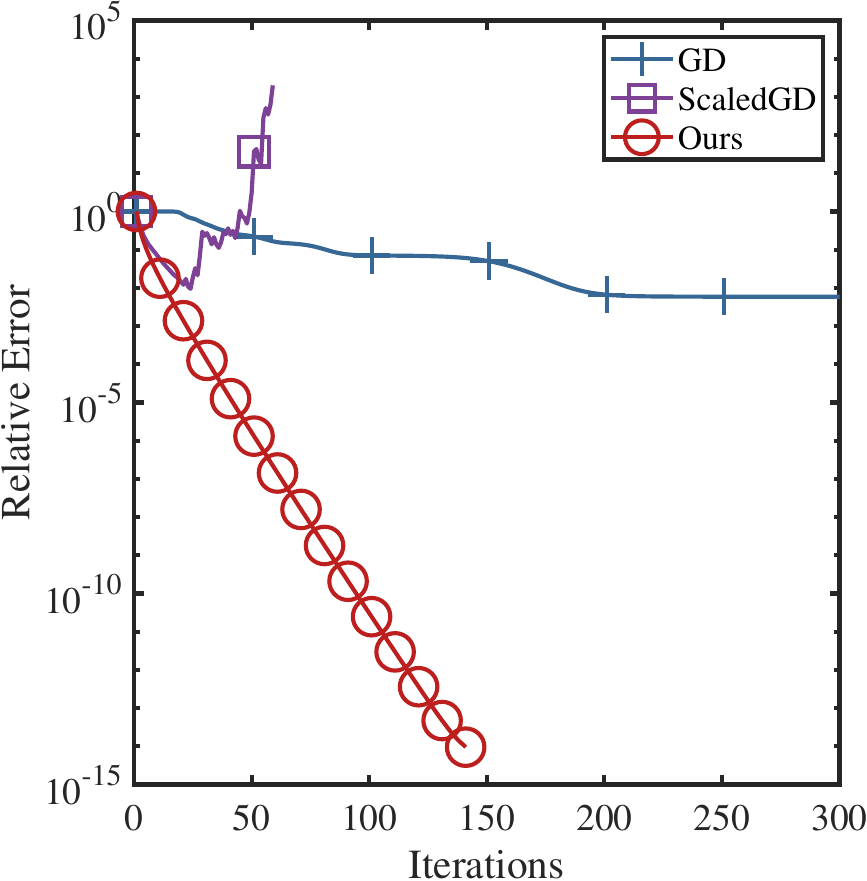}
\end{minipage}%
}
\subfigure[]{
\begin{minipage}[t]{0.48\linewidth}
\centering
\includegraphics[width=4.3cm,height=4.3cm]{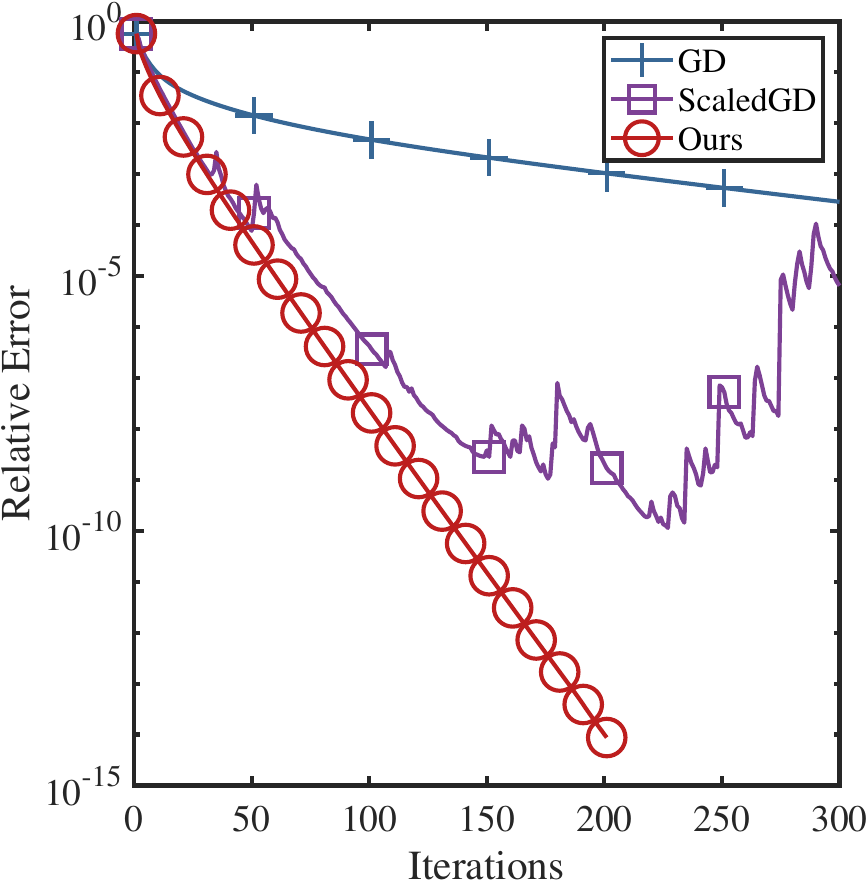}
\end{minipage}%
}
\centering
\caption{Comparison of recovery errors for GD, ScaledGD, and APGD on the low-tubal-rank tensor recovery task. Subfigure (a) shows the fully over-parameterized case with $r = 2r_\star$. Subfigure (b) uses the same tubal-rank $r = r_\star$, but $\X_\star$ is not full tubal-rank, i.e., in the Fourier domain, some frontal slices have rank smaller than $r_\star$.
}
\label{fig:1}
\end{figure}

\begin{algorithm}[tb]
\caption{Alternating Preconditioned Gradient Descent (APGD)}
\label{alg:trpca}
\textbf{Input:} Initialization $\L_0, \R_0$, estimated tubal-rank $r$, objective function $f(\L*\R^\top)$, step size $\eta$, damping parameter $\lambda_0$ 
\begin{algorithmic}[1] %[1] enables line numbers
\STATE \textbf{for} $t=0$ to $T{-1}$ \textbf{do}
\STATE \  $\widetilde{\L}_{t+\frac{1}{2}}=\L_t-\eta \nabla_{\L} f(\X_t)* (\R_t^\top *\R_t+\lambda_t \I)^{{-1}}$ 
\STATE\ $\L_{t+\frac{1}{2}},\R_{t+\frac{1}{2}} = \mathtt{Rebalance}(\tilde{\L}_{t+\frac{1}{2}},\R_t)$
\STATE\  $\widetilde{\R }_{t+1}=\R_{\small t+\frac{1}{2}}-\eta \nabla_{\R} f(\X_{\small t+\frac{1}{2}})* (\L_{\small t+\frac{1}{2}}^\top *\L_{\small t+\frac{1}{2}}+\lambda_t \I)^{{-1}}$
\STATE\ $\L_{t+1},\R_{t+1}=\mathtt{Rebalance}({\L}_{t+\frac{1}{2}},\tilde{\R}_{t+1})$
\STATE \textbf{end for}
\STATE \textbf{return:} $\X_T=\L_T*\R_T^\top$
\end{algorithmic}
\end{algorithm}

\begin{algorithm}[tb]
\caption{$\mathtt{Rebalance}$}
\label{alg:rebal}
\begin{algorithmic}[1]
\STATE \textbf{Input:}$\widetilde{\L}\in\mathbb{R}^{n_1\times r\times n_3},\widetilde{\R}\in\mathbb{R}^{n_2\times r\times n_3}$
\STATE $\Q_{\L},\W_{\L} = \text{TQR}(\widetilde{\L})$, $\Q_{\R},\W_{\R} = \text{TQR}(\widetilde{\R})$
\STATE $\hat{\U},\hat{\S},\hat{\V}=\text{TSVD}(\W_{\L}*\W_{\R}^\top)$
\STATE $\L=\Q_{\L}*\hat{\U}*\hat{\S}^{\frac{1}{2}}$, $\R=\Q_{\R}*\hat{\V}*\hat{\S}^{\frac{1}{2}}$
\STATE Output: $\L,\R$
\end{algorithmic}
\label{alg:2}
\end{algorithm}

\subsection{Our contributions}
We summarize the main contributions of this work as follows:
\begin{itemize}
    \item We observe that there exist two types of over-parameterization in the low-tubal-rank tensor estimation problem, under which previous methods such as GD and ScaledGD exhibit slow convergence or even divergence. To address this issue, we propose an Alternating Preconditioned Gradient Descent (APGD) algorithm. APGD achieves linear convergence in both over-parameterized and ill-conditioned settings. It is worth noting that this is the first general low-tubal-rank tensor estimation algorithm that achieves linear convergence in the overparameterized setting. As shown in Table 1, APGD significantly outperforms existing methods.
    \item We analyze the convergence properties of APGD and prove that, when the initial point is close to the ground truth and the objective function $f(\cdot)$ satisfies certain geometric conditions, APGD converges at a linear rate. Based on this result, we further examine several representative problems, including low-tubal-rank tensor factorization and low-tubal-rank tensor recovery, and show that our theoretical analysis applies to these cases.
    \item We conduct both synthetic experiments on the tasks of low-tubal-rank tensor factorization and low-tubal-rank tensor recovery. The experimental results demonstrate the effectiveness of APGD and show that it achieves faster convergence than existing methods such as FGD and ScaledGD. In particular, under both over-parameterized settings, GD and ScaledGD fail to converge or even diverge, while APGD consistently converges rapidly. Moreover, we conduct experiments to examine the properties of APGD and show that it is robust to the choice of the damping parameter $\lambda_t$. The experiments also demonstrate that APGD is insensitive to initialization and rebalancing, highlighting its potential for further exploration.

\end{itemize}

\begin{table}[]
\caption{Comparison of several low-tubal-rank tensor estimation methods based on t-SVD. In column ``operator", ``measurement" means the low-tubal-rank tensor recovery problem while ``projection" denotes low-tubal-rank tensor completion problem; ``over rank" denotes the over rank situation of the decomposition-based methods while ``$\backslash$" denotes that the over rank situation is not applicable to the regularization-based method. The iteration complexity is calculated on a tensor in $\mathbb{R}^{n\times n \times n_3}$ with tubal-rank $r_{\star}$ while $r$ denotes the over-parameterized tubal-rank of the decomposition-based methods.}
\centering

\begin{tabular}{cccccc}
\hline
t-SVD methods & operator &theory & rate  & over rank \\ \hline
TNN \cite{lu2018exact}  & measurement               &          \Checkmark        &        \XSolidBrush             &  $\backslash $        \\ \hline
IR-t-TNN \cite{wang2021generalized}     & measurement             &        \XSolidBrush          &       sub-linear                       &     $\backslash $      \\ \hline
RTNNM  \cite{zhang2020rip}       & measurement            &      \Checkmark           &    \XSolidBrush                              &      $\backslash $     \\ \hline
Tubal-Alt-Min \cite{liu2019low}         &   projection                  &        \Checkmark         &         \XSolidBrush                              &      \XSolidBrush     \\ \hline
TCTF \cite{zhou2017tensor}         &   projection                  &        \XSolidBrush         &         \XSolidBrush                              &      \XSolidBrush     \\ \hline
HQ-TCSAD \cite{he2023robust-b} & projection    & \XSolidBrush & \XSolidBrush  & \XSolidBrush \\ \hline

UTF\cite{du2021unifying} & projection    & \XSolidBrush & \XSolidBrush  & \Checkmark \\ \hline

FNTC\cite{jiang2023robust} & projection    & \XSolidBrush & \XSolidBrush  & \Checkmark \\ \hline 

FGD \cite{liu2024low}         &   measurement                   &    \Checkmark        &   sub-linear                                   &    \Checkmark       \\ \hline
ScaledGD \cite{wu2025guaranteed} & projection &\Checkmark & linear  &\XSolidBrush \\ \hline
Ours  & general & \Checkmark & linear  & \Checkmark \\ \hline
\end{tabular}
\label{table:1}
\end{table}
\subsection{Related work}
Early studies on low-rank tensor estimation mainly focused on convex approaches. In recent years, nonconvex methods have attracted increasing attention due to their higher computational efficiency. The related work on low-rank tensor estimation is introduced below in two parts.\\
\textbf{Non-convex methods for low-rank tensor estimation under t-SVD framework}
In recent years, non-convex approaches for solving low-tubal-rank tensor estimation problems can be broadly categorized into two groups. The first group replaces the tensor nuclear norm with non-convex surrogates, while the second group factorizes the tensor into smaller factor tensors. We begin by introducing the first class of methods. By extending the Schatten-$p$ norm, Kong et al. \cite{kong2018t} proposed a t-Schatten-$p$ norm to approximate the tensor tubal-rank. Several other non-convex surrogates inspired by t-TNN have also been proposed, such as the weighted t-TNN \cite{mu2020weighted} and the partial sum of t-TNN \cite{jiang2020multi}. Moreover, some studies replace the singular value penalty in TNN with other non-convex penalties, such as the Laplace function, to better approximate the tubal-rank \cite{cai2019tensor,xu2019laplace}.  It is worth noting that Wang et al.\cite{wang2021generalized} proposed a generalized nonconvex framework that encompasses a wide range of non-convex penalty functions with an IR-t-TNN algorithm for solving the low-tubal-rank tensor recovery problem.

However, these methods based on non-convex penalty functions focus on better approximating the tubal-rank and do not offer significant reductions in computational complexity compared to convex approaches. Therefore, decomposition-based methods are developed to improve computational efficiency. Zhou et al. \cite{zhou2017tensor} further introduced a factorization-based method with adaptive rank estimation, and established convergence to a KKT point. Based on this approach, several methods using two-factor \cite{du2021unifying,jiang2023robust} and three-factor factorizations \cite{wang2020faster,yang2025low} have been proposed to address low-tubal-rank tensor completion problems. For robust low-tubal-rank tensor completion, He and Atia \cite{he2023robust-a,he2023robust-b} proposed a  method using tensor factorization and the maximum correntropy criterion, achieving strong outlier resistance via nonconvex optimization. However, these methods lack rigorous theoretical guarantees and can only ensure convergence to a Karush–Kuhn–Tucker point. \\
\textbf{Provable non-convex low-rank tensor estimation}
Therefore, recent efforts have focused on developing non-convex low-rank tensor estimation algorithms with theoretical guarantees. Liu et al. \cite{liu2019low} proposed Tubal-Alt-Min, a nonconvex alternating minimization algorithm that factorizes a tensor into two low-tubal-rank components and solves least squares subproblems with provable convergence. For over-parameterized low-rank tensor recovery, Liu et al. \cite{liu2024low} provided local guarantees for exact recovery using factorized gradient descent. Furthermore, Karnik et al. \cite{karnik2024implicit} established the global convergence of factorized gradient descent with small random initialization and demonstrated its implicit regularization properties. However, factorized gradient descent is highly sensitive to the tensor condition number. To address this issue, Feng et al. \cite{feng2025learnable} and Wu \cite{wu2025guaranteed,wu2025fast} proposed ScaledGD for tensor RPCA and tensor completion problems, along with rigorous local convergence analysis. Recently, Wu et al. \cite{wunon,wu2025non} investigated non-convex approaches for low-tubal-rank tensor recovery under specific local sensing mechanisms, where a scaled gradient descent strategy was introduced to mitigate the slow convergence issues associated with ill-conditioned tensors. Nevertheless, ScaledGD-like methods fails in over-parameterized settings. Nevertheless, ScaledGD fails in over-parameterized settings. For other decomposition methods, numerous studies have also focused on developing non-convex algorithms with theoretical guarantees. For Tucker decomposition, Tong et al. applied ScaledGD to solve tensor RPCA \cite{dong2023fast}, robust tensor recovery \cite{tong2022accelerating}, and general low-rank tensor estimation problems \cite{tong2022scaling}, and provided theoretical guarantees. However, their analysis did not cover the over-parameterized setting. Later, Luo et al. studied the tensor-to-tensor regression problem under the over-parameterized regime \cite{luo2022tensor} and provided statistical analysis. For the tensor train decomposition, Cai et al. were the first to establish theoretical guarantees for tensor completion based on Riemannian optimization \cite{cai2022provable}. Qin et al. further provided theoretical results for nonconvex formulations of low-rank tensor recovery \cite{qin2024guaranteed}, robust tensor recovery \cite{qin2025robust}, tensor-to-tensor regression \cite{qin2025computational}, and high-order structured tensor recovery problems \cite{qin2025scalable}. From a broader theoretical perspective, Díaz et al. \cite{diaz2025preconditioned} recently established a general preconditioning framework for composite optimization, demonstrating that linear convergence can be achieved in over-parameterized regimes under mild regularity conditions.

\section{Notations and preliminaries}
\label{sec:notations}
In this paper, the terms scalar, vector, matrix, and tensor are represented by the symbols $z$, $\textbf{z}$, $\textbf{Z}$, and $\Z$ respectively. {For a 3-way tensor $\Z\in\mathbb{C}^{n_1 \times n_2 \times n_3}$, we denote its $(i,j,k)$-th entry as $\Z(i,j,k)$ and use the Matlab notation $\Z(:,:,i)$ to denote the $i$-th frontal slice. More often, the frontal slice $\Z(:,:,i)$ is denoted compactly as $\textbf{Z}^{(i)}$.} The tensor Frobenius norm is defined as $||\Z||_F = \sqrt{\sum_{i,j,k}\Z(i,j,k)^2}$. The spectral norm of matrices and tensors is denoted by $\|\cdot\|$. {For any tensor $\Z\in\mathbb{R}^{n_1 \times n_2 \times n_3}$}, $\overline{\Z}\in\mathbb{C}^{n_1\times n_2 \times n_3}$ is the Fast Fourier Transform (FFT) of $\Z$ along the third dimension. In Matlab, we have $\overline{\Z}=\mathtt{fft}(\Z,[\ ],3)$ and $\Z=\mathtt{ifft}(\overline{\Z},[\ ],3)$. {The inner product of two tensors, $\Z$ and $\Y$, is defined as $\langle \Z,\Y\rangle = \sum_{i=1}^{n_3} \langle{\bm{Z}}^{(i)},{\bm{Y}}^{(i)}\rangle$. }

\begin{definition}[Block diagonal matrix \cite{kilmer2011factorization}]
For a three-order tensor $\Z\in\mathbb{R}^{n_1\times n_2 \times n_3}$, we denote $\overline{\textbf{Z}}\in\mathbb{C}^{n_1n_3\times n_2n_3}$ as a block diagonal matrix of $\overline{\Z}$, i.e.,
$$
\overline{\bm{Z}}=\mathtt{bdiag}(\overline{\Z})=\mathtt{diag}(\overline{\bm{Z}}^{(1)};\overline{\bm{Z}}^{(2)};...;\overline{\bm{Z}}^{(n_3)}).
$$
\end{definition}

\begin{definition}[Block circulant matrix \cite{kilmer2011factorization}]
For a three-order tensor $\Z\in\mathbb{R}^{n_1\times n_2 \times n_3}$, we denote $\mathtt{bcirc}(\Z)\in\mathbb{R}^{n_1n_3\times n_2n_3}$ as its block circulant matrix, i.e.,
$$
\mathtt{bcirc(\Z)}=\begin{bmatrix}
\textbf{Z}^{(1)} & \textbf{Z}^{(n_3)} & \cdots& \textbf{Z}^{(2)}\\
\textbf{Z}^{(2)} & \textbf{Z}^{(1)} & \cdots &\textbf{Z}^{(3)}\\
\vdots & \vdots & \ddots & \vdots\\
\textbf{Z}^{(n_3)} & \textbf{Z}^{(n_3{-1})} &\cdots & \textbf{Z}^{(1)}\\
\end{bmatrix}.
$$
\end{definition}

\begin{definition}[The fold and unfold operations \cite{kilmer2011factorization}]
For a three-order tensor $\Z\in\mathbb{R}^{n_1\times n_2 \times n_3}$, we have
\begin{equation}
    \begin{aligned}
    & \mathtt{unfold}(\Z)=[\textbf{Z}^{(1)};\textbf{Z}^{(2)};\cdots;\textbf{Z}^{(n_3)}]\\
&    \mathtt{fold}(\mathtt{unfold}(\Z))=\Z.
\end{aligned}
\notag\end{equation}
\end{definition}

\begin{definition}[T-product\cite{kilmer2011factorization}]
For $\Z\in\mathbb{R}^{n_1\times n_2 \times n_3}$, $\Y\in\mathbb{R}^{n_2\times q \times n_3}$, the t-product of $\Z$ and $\Y$ is $\X\in\mathbb{R}^{n_1\times q \times n_3}$, i.e., 
\begin{equation}
\X=\Z*\Y=\mathtt{fold}(\mathtt{bcirc}(\Z)\cdot \mathtt{unfold}(\Y)).
\notag
\end{equation}
% T-product between $\A$ and $\B$ can be calculated efficiently by:\\
% 1) Calculate $\overline{\A}=\mathtt{fft}(\A,[\ ],3)$ and $\overline{\B}=\mathtt{fft}(\B,[\ ],3)$;\\
% 2) Multiply the each pair of the frontal slices of $\overline{\A}$ and $\overline{\B}$ to get $\overline\bm{\mathcal{C}}$;\\
% 3) Calculate $\mathcal{C}=\mathtt{ifft}(\overline\bm{\mathcal{C}},[\ ],3)$.
\end{definition}

\begin{definition}[Tensor conjugate transpose\cite{kilmer2011factorization}]
Let $\Z\in\mathbb{C}^{n_1\times n_2\times n_3}$, and its conjugate transpose is denoted as $\Z^{\top}\in\mathbb{C}^{n_2\times n_1\times n_3}$. The formation of $\Z^{\top}$ involves obtaining the conjugate transpose of each frontal slice of $\Z$, followed by reversing the order of transposed frontal slices 2 through $n_3$. {For an example, let $\Z\in\mathbb{C}^{n_1\times n_2\times 4}$ and its frontal slices be $\bf{Z}_1$, $\bf{Z}_2$, $\bf{Z}_3$ and $\bf{Z}_4$. Then 
$$
\Z^{\top}=\mathtt{fold}\left( \left[ \begin{array}{c}
	\bf{Z}_{1}^{\top}\\
	\bf{Z}_{4}^{\top}\\
	\bf{Z}_{3}^{\top}\\
	\bf{Z}_{2}^{\top}\\
\end{array} \right] \right),
$$
where $\bm{Z}^{\top}\in\mathbb{C}^{n_2 \times n_1}$ denotes the matrix conjugate transpose of  $\bm{Z}\in\mathbb{C}^{n_1 \times n_2}$.}
\end{definition}

\begin{definition}[Identity tensor\cite{kilmer2011factorization}]
 The identity tensor, represented by $\I\in\mathbb{R}^{n\times n\times n_3}$, is defined such that its first frontal slice corresponds to the $n\times n$ identity matrix, while all subsequent frontal slices are comprised entirely of zeros. This can be expressed mathematically as:
\begin{equation}
\I^{(1)} = I_{n\times n},\quad \I^{(i)} = 0,i=2,3,\ldots,n_3.
\notag\end{equation}
\end{definition}

\begin{definition}[Orthogonal tensor \cite{kilmer2011factorization}]
A tensor $\Q\in\mathbb R^{n\times n\times n_3}$ is considered orthogonal if it satisfies the following condition:
\begin{equation}
\Q^{\top} * \Q = \Q * \Q^{\top}=\I.
\notag\end{equation}
\end{definition}

\begin{definition}[F-diagonal tensor \cite{kilmer2011factorization}]
A tensor is called f-diagonal if each of its frontal slices is a diagonal matrix.
\end{definition}

\begin{theorem}[t-SVD \cite{kilmer2011factorization,lu2018exact}]
Let $\Z\in\mathbb{R}^{n_1\times n_2 \times n_3}$, then it can be factored as 
\begin{equation}
    \Z=\U * \S * \V^{\top},
\notag\end{equation}
where $\U\in\mathbb{R}^{n_1 \times n_1 \times n_3}$, $\V\in\mathbb{R}^{n_2\times n_2 \times n_3}$ are orthogonal tensors, and $\S\in\mathbb{R}^{n_1\times n_2 \times n_3}$ is a f-diagonal tensor.
\end{theorem}

\begin{definition}[Tubal-rank \cite{kilmer2011factorization}]
For $\Z\in\mathbb{R}^{n_1\times n_2 \times n_3}$, its tubal-rank as rank$_t(\Z)$ is defined as the nonzero {diagonal} tubes of $\S$, where $\S$ is the f-diagonal tensor from the t-SVD of $\Z$. That is
\begin{equation}
\operatorname{rank}_t(\Z):= \#\{i:S(i,i,:)\neq 0\}.
\notag\end{equation}
\end{definition}

\begin{definition}
[Multi-rank \cite{kilmer2021tensor}]
For $\X\in\mathbb{R}^{n_1\times n_2 \times n_3}$ with tubal-rank $r$, its multi-rank is a vector $\r_m\in\mathbb{R}^{n_3}$, i.e., rank$_m(\X)=\r_m$, with its $i$-th  entry being the rank of the $i$-th frontal slice $\overline{\bm{X}}^{(i)}$ of $\overline{\X}$, i.e., $r_i=$rank$(\overline{\bm{X}}^{(i)})$. Obviously, the tubal-rank $r$ of $\X$ is the max entry of the multi-rank vector, i.e., $r=\max(\r_m)$. And we denote $s_{r}^m=||\r_m||_1$ for notation convenience.
\end{definition}

\begin{definition}[Tensor singular value]
For any tensor $\X\in\mathbb{R}^{n_1\times n_2\times n_3}$ with tubal-rank $r$ and multi-rank $\r_m$, its singular values are defined as
$$
\sigma_i (\X) = \sigma_i(\overline{\bf{X}}),\ i=1,..., ||\r_m||_1,
$$
where {$\overline{\bm{X}}$ is the block diagonal matrix of tensor $\overline{\X}$ and }$\sigma_1(\overline{\bm{X}})\ \ge \sigma_2(\overline{\bm{X}})\ge ...\ge \sigma_{||\r_m||_1}(\overline{\bm{X}})>0 $ denote the singular values of $\overline{\bm{X}}$. The condition number of $\X$ is defined as 
$$
\kappa(\X):=\frac{\sigma_1(\overline{\bm{X}})}{\sigma_{||\r_m||_1}(\overline{\bm{X}})}.
$$
\end{definition}

\begin{definition}[Tensor spectral norm \cite{lu2018exact}]
For $\Z\in\mathbb{R}^{n_1 \times n_2 \times n_3}$, its spectral norm is denoted as 
\begin{equation}
    \|\Z\|=\|\mathtt{bcirc}(\Z)\|=\|\overline{\bm{Z}}\|.
\notag\end{equation}
\end{definition}

\section{Main results}
\label{sec:main results}
First, we establish the linear convergence guarantee of APGD under general geometric assumptions on the loss function $f(\cdot)$. Then, based on this theory, we analyze two representative problems, the low-tubal-rank factorization problem and the low-tubal-rank tensor recovery problem.

\subsection{Problem formulation and assumptions}
Let $\X_\star\in\mathbb{R}^{n_1\times n_2\times n_3}$ be the ground truth tensor with tubal-rank $r_\star$ and multi-rank $\r_m^\star\in\mathbb{R}^{n_3}$, then it can be factorized as $\L_\star *\R_\star ^\top$, where $\L_\star\in\mathbb{R}^{n_1\times r_\star \times n_3 },\ \R_\star\in\mathbb{R}^{n_2\times r_\star \times n_3}$. Define the tensor singular value decomposition as $\X_\star=\U_\star*\S_\star*\V_\star^\top$, and the condition number of $\X_\star$ is defined as $\kappa:={\sigma_1(\overline{\bm{X}_\star})}/{\sigma_{\min}(\overline{\bm{X}_\star})}.$

We then introduce two geometric properties of the loss function $f(\cdot)$, which play a crucial role in our subsequent analysis.

\begin{definition}[Tensor $(L,r)$-restricted smoothness]
A differentiable function $\Phi: \mathbb{R}^{n_1 \times n_2 \times n_3}\mapsto \mathbb{R}$ is said to be $(L,r)$-restricted smooth for some $L>0$ if
$$
f(\X_2) \le f(\X_1) + \langle \nabla f(\X_1), \X_2-\X_1 \rangle + \frac{L}{2}\|\X_2 -\X_1\|_F^2
$$
for any $\X_1,\ \X_2\in\mathbb{R}^{n_1 \times n_2 \times n_3}$ with tubal-rank at most $r$.
\end{definition}

\begin{definition}[Tensor $(\mu,r)$-restricted strong convexity]
A differentiable function $f: \mathbb{R}^{n_1 \times n_2 \times n_3}\mapsto \mathbb{R}$ is said to be $(\mu,r)$-restricted convex for some $\mu\ge 0$ if 
$$
f(\X_2) \ge f(\X_1) + \langle \nabla f(\X_1), \X_2-\X_1 \rangle +\frac{\mu}{2} \| \X_2 - \X_1 \|_F^2,
$$
for any $\X_1,\ \X_2\in\mathbb{R}^{n_1 \times n_2 \times n_3}$ with tubal-rank at most $r$.    
\end{definition}

Based on these two definitions, we state the following lemma.
\begin{lemma}
Suppose $f(\X)$ is $(L,2r)$-restricted smooth and $(\mu,2r)$-restricted strongly convex. Let $\X_\star = \operatorname{arg min}f$ satisfy $\operatorname{rank}_t(\X_\star)=r_\star\le r$. Then we have 
$$
\frac{\mu}{2} \| \X - \X_\star \|_F^2 \le f(\X) -f(\X_\star) \le \frac{L}{2} \| \X - \X_\star \|_F^2.
$$
\label{lemma:1.3}
\end{lemma}

\begin{IEEEproof}
This lemma can be directly derived from the definitions of ($L,2r$)-restricted smoothness and $(\mu,2r)$-strong convexity.
\end{IEEEproof}
This lemma establishes the relationship between the loss function and the recovery error, allowing us to derive an upper bound on $||\X - \X_\star||_F^2$ by analyzing $f(\X) - f(\X_\star)$.

Encouragingly, many low-tubal-rank tensor estimation problems satisfy the two set-based assumptions stated above, including but not limited to:
\begin{itemize}
    \item \textit{low-rank matrix estimation:} The low-tubal-rank tensor estimation problem (\ref{equ:03}) naturally reduces to various classes of low-rank matrix estimation problems, including but not limited to weighted PCA \cite{srebro2003weighted}, matrix sensing \cite{zhuo2021computational}, and quadratic sampling \cite{li2019nonconvex}. For detailed analysis, please refer to \cite{tong2022accelerating,li2019non}
    \item \textit{low-tubal-rank tensor factorization:} The loss function for low-tubal-rank tensor factorization is $f(\X) = \frac{1}{2}||\X-\X_\star||_F^2$, where rank$_t(\X_\star)=r_\star$ and rank$_t(\X) = r$. It is straightforward to see that $f(\X)$ satisfies $(L,2r)$-smoothness and $(\mu,2r)$-restricted strong convexity with $L = \mu = 1$.
    \item \textit{low-tubal-rank tensor recovery:} The loss function for low-tubal-rank tensor recovery is $$f(\X) = \frac{1}{2}||\M(\X-\X_\star)||_2^2,$$ where rank$_t(\X_\star)=r_\star$ and rank$_t(\X) = r$. It is straightforward to show that $f(\X)$ satisfies $(L,2r)$-restricted smoothness and $(\mu,2r)$-restricted strong convexity with $L= 1+\delta,\ \mu=1-\delta$, provided the linear map $\M$ satisfies the T-RIP property; see Section \ref{sec:LTRTR} for details.

    \item \textit{1-bit low-tubal-rank tensor recovery:} The loss function for 1-bit low-tubal-rank tensor recovery is $$f(\X)=\sum_{i=1}^{n_1}\sum_{j=1}^{n_2}\sum_{k=1}^{n_3} \left( \log(1+e^{\X_{ijk}}) -\alpha_{ijk}\X_{ijk} \right),$$ where $\alpha_{ijk}\in\{0,1\}$ is the observed 1-bit label for entry $(i,j,k)$ and $\mathbb{P}(\alpha_{ijk}=1|\X_{ijk})=\sigma(\X_{ijk})$, with $\sigma(\cdot)$ denoting the sigmoid function. Under the common boundedness assumption $\max(|\X_{ijk}|) \le c_t$, one can check that $f(\X)$ satisfies $(L,2r)$-restricted smoothness and $(\mu,2r)$-restricted strong convexity with $L=1/4$ and $\mu={\sigma(c_t)}{(1-\sigma(c_t))}.$
\end{itemize}

Beyond the geometric assumptions, we also require the initial point to be sufficiently close to the ground truth, specified as follows:
\begin{assumption}[\textbf{Initialization Assumption}]
The initial point $\X_0$ satisfies $||\X_0-\X_\star||_F\le\rho \sigma_{\min}(\bar{\bf{X}}_\star),\ \rho=\sqrt{\frac{1}{(1+4\kappa L/\mu)n_3}}$.
\end{assumption}
This initialization condition is a common assumption in prior work and can be achieved through spectral initialization \cite{wu2025guaranteed,feng2025learnable}. The specific form of spectral initialization varies across different problems, which we will describe in Section \ref{sec:4}.

\subsection{Main theorem}
With the aforementioned assumptions, we present the main result directly.
\begin{theorem}
Suppose that \( f \) satisfies the \((L,2r)\)-restricted smoothness and $(\mu,2r)$-restricted strongly convexity. Then, for the low-tubal-rank tensor estimation problem (\ref{equ:03}), starting from an initial point $\X_0\in\mathbb{R}^{n_1\times n_2\times n_3}$ with tubal-rank $r\ge r_\star$ and multi-rank $\r_m=[r,...,r]\in\mathbb{R}^{n_3  }$ that satisfies $||\X_0-\X_\star||_F\le\rho\sigma_{\min}(\bar{\bf{X}}_\star),\ \rho=\sqrt{\frac{1}{(1+4\kappa L/\mu)n_3}}$, solving it using APGD with step size $\eta \le \frac{1}{L} $ and $\lambda_t\le  \sqrt{\frac{2}{Lc_1^2}}(f(\X_t)-f(\X_\star))^{1/2}$ leads to
\begin{equation}
\begin{aligned}
f(\X_{t}) - f(\X_\star) &\le (1-q)^{2t}\left[ f(\X_0)-f(\X_\star) \right],\\
||\X_t-\X_\star ||_F^2 & \le \frac{\mu}{L}(1-q)^{2t} ||\X_0-\X_\star||_F^2,
\end{aligned}
\notag
\end{equation}
where $q=(\eta -\frac{L\eta^2}{2})\frac{\mu^2}{6L}$, $c_1=\left(\frac{1}{\sqrt{5}-1}+ \frac{\sqrt{2(s_r^m-s_{r_\star}^m)}(L+\mu)}{\sqrt{\mu Ln_3}}\right)$, $s_r^m=||\r_m||_1$, and $s_{r_\star}^m=||\r_\star||_1.$
\label{theorem:main}
\end{theorem}
\begin{IEEEproof}
See Section \ref{sec:proof sketch}.
\end{IEEEproof}

\begin{remark}[\textbf{Linear convergence rate}]
This theorem shows that when the objective function satisfies certain geometric conditions and the initialization is sufficiently close to the ground truth, APGD achieves linear convergence. Moreover, its convergence rate is independent of the condition number of the target tensor $\X_\star$, even in the over-parameterized setting. Notably, to the best of our knowledge, this is the first non-convex tensor estimation framework that achieves linear convergence under over-parameterization.
\end{remark}

\begin{remark}[\textbf{Choice of damping parameter $\lambda$}]
The assumption on $\lambda$ in Theorem 1 is $\lambda_t\le  \sqrt{\frac{2}{Lc_1^2}}[f(\X_t)-f(\X_\star)]^{1/2}$, which ties $\lambda_t$ to the per-iteration error. We may also choose a very small fixed value. For example, if the final recovery error satisfies $|| \X_T-\X_\star ||_F = \Theta(\epsilon)$, setting $\lambda \le \epsilon / c_1$ still yields the result of Theorem 2; see Section \ref{sec:proof sketch} for details. Hence, APGD is robust to the choice of the damping parameter.
\end{remark}

\begin{remark}[\textbf{Comparison with ScaledGD \cite{wu2025guaranteed}}]
Compared with ScaledGD, the main advantage of APGD is its ability to handle over-parameterized cases. The key difference lies in the preconditioning terms $(\L^\top * \L)^{{-1}}$ and $(\R^\top * \R)^{{-1}}$, where APGD introduces a damping term $\lambda \I$. This ensures that $\L^\top * \L + \lambda \I$ and $\R^\top * \R + \lambda \I$ are invertible. However, adding the damping term $\lambda \I$ breaks the covariance property of the updates for $\L$ and $\R$, which is a crucial property in ScaledGD. As a result, the theoretical analysis of APGD differs substantially from that of ScaledGD. Furthermore, due to the use of alternating updates, APGD is more robust to the choice of step size and can converge with a larger step size.
\end{remark}

\begin{remark}[\textbf{Comparison with PrecGD \cite{zhang2023preconditioned} in matrix case}]
In low-rank matrix estimation, Zhang et al. \cite{zhang2023preconditioned} proposed PrecGD to extend ScaledGD to the over-parameterized setting. Similarly, PrecGD introduces a damping term $\lambda \I$ to prevent the preconditioning matrices from becoming singular. There are three key differences between APGD and PrecGD.
First, PrecGD focuses on the symmetric case and thus does not need to address the imbalance between the two factors, which simplifies the analysis.
Second, PrecGD imposes strict upper and lower bounds on the damping parameter.
Finally, PrecGD is limited to matrices, whereas APGD addresses the over-parameterized tensor setting, which involves more intricate tensor-specific analysis and is therefore considerably more complex.
\end{remark}

\begin{remark}[\textbf{Technique challenges}]
At first glance, APGD appears to differ from ScaledGD only by the addition of a damping term $(\lambda \I)$ in the scaling. However, under over-parameterization, the analytical framework becomes fundamentally different. First, over-parameterization destroys the invertibility and covariance properties of ScaledGD. Second, there exist two distinct forms of over-parameterization, which render previous tubal-rank-based analyses invalid. To address this, we first apply a rebalancing step to align the factor tensors and decompose the update into two preconditioned subproblems (for $\L$ and $\R$) that are analyzed separately. In addition, we utilize the tensor multi-rank to capture tensor low-rank structure at a finer granularity, thereby unifying the two types of over-parameterization within a single analytical framework and completing the convergence proof.
\end{remark}

\begin{remark}\textbf{(Computational complexity)}
The main computational cost of APGD comes from four components:
\begin{itemize}
    \item the cost of computing the tensor product involving the factor tensors is $\mathcal{O}\left( rn_3(n_1\lor n_2) \log n_3+ r(n_1\lor n_2)^2n_3\right )$;
    \item the cost of inverting an $r \times r \times n_3$ tensor is $\mathcal{O}(r^3n_3)$;
    \item the cost of performing a tensor QR decomposition on the factor tensors $\L\in\mathbb{R}^{n_1\times r\times n_3}$ and $\R\in\mathbb{R}^{n_2\times r \times n_3}$ in Algorithm \ref{alg:2}  is $\mathcal{O}\left(r^2(n_1\lor n_2)n_3\right)$;
    \item the cost of computing a t-SVD of an $r \times r \times n_3$ tensor in Algorithm \ref{alg:2} is $\mathcal{O}(r^3n_3)$.
\end{itemize}
Combining these four parts, the overall computational complexity is $\mathcal{O}\left ( r(n_1\lor n_2)^2n_3 +  rn_3(n_1\lor n_2) \log n_3 \right)$, where we assume that $r\ll (n_1\land n_2)$. For the traditional tensor nuclear norm minimization method, the computational complexity is $\mathcal{O}(n_1n_2n_3\log n_3+ (n_1\lor n_2 )(n_1\land n_2)^2n_3)$. For ScaledGD, its complexity is $\mathcal{O}(r(n_1\lor n_2)^2n_3 +  rn_3(n_1\lor n_2) \log n_3)$, which is of the same order of APGD. For FGD, its complexity is also $\mathcal{O}(r(n_1\lor n_2)^2n_3 +  rn_3(n_1\lor n_2) \log n_3).$ Comparing these methods, we observe that when $r \ll (n_1 \lor n_2) $, the computational complexities of APGD, ScaledGD, and FGD are of the same order and significantly lower than that of TNN. However, in practice, each iteration of APGD requires updating the gradients twice and performing two rebalancing operations, making the computational time per iteration slightly higher than FGD and ScaledGD. Nevertheless, since APGD is more robust to the step size, we can actually choose a larger step size, which reduces the total computation time. We will validate this in the experiments presented later.

\end{remark}

\subsection{ Why ScaledGD fails in the over-parameterized case}
We first analyze the problem of ScaledGD and explain why ScaledGD fails in the over-parameterized setting. ScaledGD can be viewed as a gradient descent method under a \textbf{new metric}. Within the framework of ScaledGD, the $\P$-norm, and dual $\P$-norm are defined by
\begin{equation}
    \begin{aligned}
&||\D||_P :=\left\| \begin{bmatrix}
    \D_1*(\R^\top\R)^{\frac{1}{2}} \\ 
    \D_2*(\L^\top\L)^{\frac{1}{2}}
\end{bmatrix}  \right\|_F   \\  &||\D||_{P^*} :=\left\| \begin{bmatrix}
    \D_1*(\R^\top\R)^{-\frac{1}{2}} \\ 
    \D_2*(\L^\top\L)^{-\frac{1}{2}}
\end{bmatrix}  \right\|_F,
\end{aligned}
\label{equ:05}
\end{equation}
where 
$$
\D_1=\nabla_{\L}f(\X)*(\R^\top *\R)^{-1},\ \D_2=\nabla_{\R}f(\X)*(\L^\top *\L)^{-1}.
$$

For the factorization form $\X=\L*\R^\top$, define $g(\Z):=f(\L*\R^\top)$, where
\begin{align*}
    \Z:=\begin{bmatrix}
    \L \\ \R
\end{bmatrix} \in\mathbb{R}^{(n_1+  n_2)\times r \times n_3}, \     \D:=\begin{bmatrix}
    \D_1 \\ \D_2
\end{bmatrix} \in\mathbb{R}^{(n_1+  n_2)\times r \times n_3}.
\end{align*}

Based on the above conditions, we derive a Lipschitz-type inequality under the new metric for the simple low-tubal-rank tensor factorization.
\begin{lemma}
For $g(\Z):=f(\X)=\frac{1}{2}||\X-\X_\star||_F^2$,
let $||\D||_P$ defined as Equation (\ref{equ:05}), then minimize $f(\X)$ via ScaledGD leads to
\begin{equation}
g(\Z+\D) \le g(\Z) + \langle \nabla g(\Z) ,\D \rangle +\frac{ L_P}{2} ||\D||_P^2,
\notag
\end{equation}
where $L_P=2+ \frac{||\D||_P^2}{2\sigma^4_{\min}} + \frac{\sqrt{2}||\E||_F}{\sigma^2_{\min}} + \frac{2||\D||_P}{\sigma^2_{\min}},$
where $\sigma_{\min}=\min(\sigma_{\min}(\bar{\bf{L}}),\sigma_{\min}(\overline{\bf{R}})).$
\label{lemma:2}
\end{lemma}
\begin{IEEEproof}
See Section II  of the supplementary material.
\end{IEEEproof}

From this lemma, we see that under the new metric, the Lipschitz constant $L_P$ of ScaledGD is inversely proportional to $\sigma_{\min}^2$. In the exact-rank and full tubal-rank case, as $ \X_t \to \X_\star $, we have $\sigma_{\min}(\bar{\bf{L}}_t  \bar{\bf{R}}_t^\top) \to \sigma_{\min}(\bar{\bf{X}}_\star)$, so $\sigma_{\min}$ does not become too small. However, in the over-parameterized regime, $\sigma_{\min}(\bar{\bf{L}}_t  \bar{\bf{R}}_t^\top) \to 0$. Consequently, $L_P \to \infty$, which causes ScaledGD to break down, as shown in Figure \ref{fig:1}.

\subsection{Proof sketch}
\label{sec:proof sketch}

Therefore, a reasonable approach is to add a damping term $\lambda \I$. Accordingly, we define new local norms: 
\begin{equation}
\begin{aligned}
&||\D_1||_{P_R}:=||\D_1*(\R^\top*\R+\lambda \I)^{\frac{1}{2}}||_F,\\ &||\D_1||_{P^*_R}:=||\D_1*(\R^\top*\R+\lambda \I)^{-\frac{1}{2}}||_F,\\
&||\D_2||_{P_L}:=||\D_2*(\L^\top*\L+\lambda \I)^{\frac{1}{2}}||_F,\\ &||\D_2||_{P^*_L}:=||\D_2*(\L^\top*\L+\lambda \I)^{-\frac{1}{2}}||_F.
\end{aligned}
\end{equation}
Based on these new metrics, we derive a corresponding Lipschitz-type inequality.

\begin{lemma}
    Suppose that $f$ satisfies the $(L,2r)$-restricted smoothness, then solving the low-tubal-rank tensor estimation problem via APGD leads to
\begin{align*}
     f\left(\X_{t+\frac{1}{2}}\right) &\le f(\X_t) -\eta \langle \nabla_{\L}f(\X_t), \D_{\L}^t \rangle +\frac{\eta^2L}{2}|| \D_{\L}^t ||_{P_R}^2\\
     f(\X_{t+1}) &\le f\left(\X_{t+\frac{1}{2}}\right) -\eta \langle \nabla_{\R} f\left(\X_{t+\frac{1}{2}}\right),\D_{\R}^t \rangle \\
    & + \frac{\eta^2 L}{2} ||\D_{\R}^t||_{P_L}^2.
\end{align*}
\label{lemma:03}
\end{lemma}
\begin{IEEEproof}
Note that 

\begin{align*}
&f\left(\X_{t+\frac{1}{2}}\right) = f((\L_t-\eta \D_{\L}^t)*\R_t^\top) = f(\X_t-\eta \D_{\L}^t*\R_t^\top)\\
    &\overset{(a)}{\le} f(\X_t) -\eta \langle \nabla f(\X_t),\D_{\L}^t*\R_t^\top \rangle + \frac{\eta^2 L}{2} ||\D_{\L}^t*\R_t^\top||_F^2\\
    &=f(\X_t) -\eta \langle \nabla f(\X_t)*\R_t,\D_{\L}^t \rangle +\\
    & \frac{\eta^2 L}{2} ||\D_{\L}^t{\small *}[(\R_t^\top *\R_t +\lambda \I)^{\frac{1}{2}}\small{*}(\R_t^\top *\R_t+ \lambda \I)^{\small \tiny{-\frac{1}{2}}}]*\R_t^\top||_F^2\\
    &\overset{(b)}{\le} f(\X_t) -\eta \langle \nabla_{\L}f(\X_t), \D_{\L}^t \rangle +\frac{\eta^2L}{2}|| \D_{\L}^t ||_{P_R}^2,
\end{align*}
where (a) uses the $(L,2r)$-restricted smoothness; (b) uses the facts that $||\A*\B||_F\le ||\A||_F||\B||$ and $$||(\R_t^\top *\R_t +\lambda \I)^{-\frac{1}{2}}*\R_t^\top||\le 1.$$
Similarly, we can obtain
$$
 f(\X_{t+1}) \le f(\X_{t+\frac{1}{2}}) -\eta \langle \nabla_{\R} f(\X_{t+\frac{1}{2}}),\D_{\R}^t \rangle + \frac{\eta^2 L}{2} ||\D_{\R}^t||_{P_L}^2.
$$
Therefore, we complete the proof of Lemma \ref{lemma:03}.
\end{IEEEproof}

\begin{remark}
From this lemma, we can see that the Lipschitz constant of APGD is a fixed value and does not change during the iterations, which is a significant difference from ScaledGD. This is because we adopt an alternating update scheme, which splits the non-convex optimization problem into two subproblems and thus avoids the influence of cross terms of two factors. Therefore, APGD can still converge even in the over-parameterized case. Moreover, the step size constraint for APGD is $\eta \le \frac{1}{L}$, while for ScaledGD it is $\eta \le \frac{1}{L_P}$. 
\end{remark}

Note that using APGD, we have 
\begin{align*}
&||\D_{\L}^t||_{P_R} \\
&\quad= ||\nabla_{\L}f(\X_t)*(\R_t^\top *\R_t + \lambda \I)^{-\frac{1}{2}}||_F=||\nabla_{\L}f(\X_t)||_{P_R^*}\\
&||\D_{\R}^t||_{P_L}\\
&\quad = ||\nabla_{\R}f(\X_t)*(\L_t^\top *\L_t + \lambda \I)^{-1}||_F=||\nabla_{\R}f(\X_t)||_{P_L^*}.
\end{align*}

If we can establish lower bounds for these two terms $||\D_{\L}^t||_{P_R}$ and $||\D_{\R}^t||_{P_L}$, similar to a PL-type inequality, namely 
\begin{equation}
\begin{aligned}
&||\nabla_{\L}f(\X_t)||_{P_R^*} \ge \mu_p [f(\X_t)-f(\X_\star)]\\ &||\nabla_{\L}f\left(\X_{t+\frac{1}{2}}\right)||_{P_R^*} \ge \mu_p [f\left(\X_{t+\frac{1}{2}}\right)-f(\X_\star)], 
\label{equ:007}
\end{aligned}
\end{equation}
then we can proceed further. Combining this with Lemma 3, we can obtain the proof of Theorem \ref{theorem:main}.

% \begin{lemma}[Descent lemma]
%     Suppose that $f$ satisfies the $(L,2r)$-restricted smoothness, then solving the low-tubal-rank tensor estimation problem via APGD leads to
%     \begin{equation}
%     \begin{aligned}
%         &f\left(\X_{t+\frac{1}{2}}\right) - f(\X^\star)\le f(\X_t)- f(\X^\star) \\
% &\ \ \ \ -(\eta -\frac{L\eta^2}{2}) \| \nabla_{\L} f(\X_t) * (\R_t^\top * \R_t + \lambda \I)^{-\frac{1}{2}} \|_F^2,\\
% &f(\X_{t+1}) - f(\X^\star)\le f\left(\X_{t+\frac{1}{2}}\right)- f(\X^\star) \\
% &\ \ \ \  -(\eta -\frac{L\eta^2}{2}) \| \nabla_{\R} f\left(\X_{t+\frac{1}{2}}\right) * (\L_{t+1}^\top * \L_{t+1} + \lambda \I)^{-\frac{1}{2}} \|_F^2.
%     \end{aligned}
%     \notag
%     \end{equation}
% \label{lemma:2}
% \end{lemma}

To obtain such lower bounds, we proceed in three steps.
Firstly, we expand the gradient norm in terms of the alignment between the error term $\E$ and the top-$k$ tangent space, as shown in Lemma \ref{lemma:011}.
\begin{lemma}
Assume that $f$ is $L$-smooth and $(\mu,\ 2r)$-restricted strongly convex, then we have
\begin{align*}
&\frac{||\nabla_{\L}f(\L*\R_t^\top)||_{P_R^*}}{||\L*\R^\top-\X_\star||_F}\ge \underset{k\in\{1,..,s_r^m\}}{\max} \frac{\mu+L}{2}    \frac{\cos\vartheta_{\R}^k-\varepsilon}{\sqrt{1+\lambda/\sigma_{k}^2(\bar{\bf{R}})}}\\
&\frac{||\nabla_{\R}f(\L*\R_t^\top)||_{P_L^*}}{||\L*\R^\top-\X_\star||_F}\ge \underset{k\in\{1,..,s_r^m\}}{\max} \frac{\mu+L}{2}  \frac{\cos\vartheta_{\L}^k-\varepsilon}{\sqrt{1+\lambda/\sigma_{k}^2(\bar{\bf{L}})}},
\end{align*}
where $\varepsilon=\frac{L-\mu}{L+\mu}$ and $\cos\vartheta_{\R}^k,\ \cos\vartheta_{\L}^k$ are defined as
\begin{align*}
&\cos\vartheta_{\R}^k= \max_{\O_1\in\mathbb{R}^{n_1\times r\times n_3}} \frac{\langle \bf{\bar{E}},\bf{\bar{O}_1} \bar{\bf{R}}_k\rangle}{||\bar{\bf{E}}||_F ||\bar{\bf{O}}_1\bar{\bf{R}}^\top_k||_F},\\
&\cos\vartheta_{\L}^k= \max_{\O_2\in\mathbb{R}^{n_2\times r\times n_3}} \frac{\langle \bar{\bf{E}},\bar{\bf{L}}\bar{\bf{O}}_2^\top \rangle}{||\bar{\bf{E}}||_F ||\bar{\bf{O}}_2^\top\bar{\bf{L}}_k||_F},
\end{align*}
and $\bar{\bf{R}}_k$ and $\bar{\bf{L}}_k$ denote the top-$k$ SVD truncation of $\bar{\bf{R}}$ and  $\bar{\bf{L}}$, and $s_r^m=||\r_m||_1$.
\label{lemma:011}
\end{lemma}
\begin{IEEEproof}
See Section III of the supplementary material.
\end{IEEEproof}

Secondly, we show that when the initialization is close to the ground truth, the principal angles between the tensor column and row spaces are small, as shown in Lemma \ref{lemma:25}.
\begin{lemma}
Suppose that the initialization satisfies $||\X_0-\X_\star||_F\le\rho\sigma_{\min}(\bar{\bf{X}}_\star),\rho=\sqrt{\frac{1}{(1+4\kappa L/\mu)n_3}}$, then we have  
\begin{align*}
    \sin\theta_{\L}^t &:= \frac{||(\I-\P^t_{\L})*\X_\star||_F}{||\L_t*\R_t^\top -\X_\star||_F} \le \frac{\mu}{4L}\\
    \sin\theta_{\R}^t &:= \frac{||\X_\star*(\I-\P^t_{\R})||_F}{||\L_t*\R_t^\top -\X_\star||_F} \le \frac{\mu}{4L},
\end{align*}
where \begin{align*}
    \P_{\L}^t=\L_t*(\L_t^\top * \L_t)^{\dagger}*\L_t^\top,
    \P^t_{\R}=\R_t*(\R_t^\top *\R_t)^{\dagger}*\R_t^\top.
\end{align*}
\label{lemma:25}
\end{lemma}
\begin{IEEEproof}
See Section IV of the supplementary material.
\end{IEEEproof}
Thirdly, combining the first two steps and using induction, we prove that if $\E$ is not well aligned with the top-$k$ tangent space, its alignment with the $(k+1)$-th tangent space becomes better, and this continues until the alignment is sufficient. Then we get the PL-type inequality, as shown in Lemma \ref{lemma:3}.
\begin{lemma}
Suppose that $f$ satisfies the $(L,2r)$-restricted smoothness and $(\mu,2r)$-restricted strong convexity, and the initialization satisfies $||\X-\X_\star||_F\le \rho \sigma_{\min}(\bar{\bf{X}}_\star),\rho=\sqrt{\frac{1}{(1+4\kappa L/\mu)n_3}}$, then solving the low-tubal-rank tensor estimation problem via APGD leads to
\begin{equation}
\begin{aligned}
& \frac{||\nabla_{\L}f(\L_t*\R_t^\top)||_{P_R^*}}{[f(\X_t)-f(\X_\star)]^{\frac{1}{2}}} \ge \sqrt{\frac{\mu^2}{2L}} \left(2+ c_1 \frac{\lambda }{||\X_t-\X_\star||_F}   \right)^{-\frac{1}{2}} \\ 
&\frac{||\nabla_{\R}f(\L_t*\R_t^\top)||_{P_L^*}}{[f(\X)-f(\X_\star)]^{\frac{1}{2}}} \ge \sqrt{\frac{\mu^2}{2L}}  \left(2+ c_1 \frac{\lambda }{||\X_t-\X_\star||_F}   \right)^{-\frac{1}{2}} ,
\end{aligned}
\notag
 \end{equation}
where $c_1=\left(\frac{1}{\sqrt{5}-1}+ \frac{\sqrt{2(s_r^m-s_{r_\star}^m)}(L+\mu)}{\sqrt{\mu Ln_3}}\right).$
\label{lemma:3}
\end{lemma}
\begin{IEEEproof}
See Section V of the supplementary material.
\end{IEEEproof}
From Lemma 6, we can see that a key factor for establishing a PL-type inequality is choosing an appropriate damping parameter $\lambda$. When we set $\lambda_t \le ||\X_t - \X_\star||_F / c_1$, we have 
\begin{equation}
\begin{aligned}
&\frac{||\nabla_{\L}f(\L_t*\R_t^\top)||_{P_R^*}}{[f(\X_t)-f(\X_\star)]^{\frac{1}{2}}} \ge \sqrt{\frac{\mu^2}{6L}},\\ &\frac{||\nabla_{\R}f(\L_t*\R_t^\top)||_{P_L^*}}{[f(\X)-f(\X_\star)]^{\frac{1}{2}}} \ge \sqrt{\frac{\mu^2}{6L}}.
\end{aligned}
\label{equ:008}
\end{equation}
Alternatively, $\lambda$ can be set to a very small fixed value. For example, if the final error satisfies $||\E_{\mathrm{final}}||_F = \Theta(\epsilon)$, then choosing $\lambda \le \epsilon / c_1$ is sufficient.

By combining the Lemmas \ref{lemma:03} and \ref{lemma:3}, a few straightforward derivation yield Theorem \ref{theorem:main}.

\section{Analysis of stylized applications}
\label{sec:4}
\subsection{Low-tubal-rank tensor factorization}
 We first consider the application of APGD in the low-tubal-rank tensor factorization problem, which can be considered as a high-order extension of the low-rank matrix factorization problem \cite{ye2021global,jiang2023algorithmic}. This problem serves as the foundation for many low-tubal-rank tensor estimation tasks, such as low-tubal-rank tensor completion, tubal tensor RPCA, and low-tubal-rank tensor recovery.
 Based on the tensor BM factorization, the objective function of low-tubal-rank tensor factorization can be written as 
\begin{equation}
    f(\X):=\frac{1}{2} ||\X-\X_\star||_F^2=\frac{1}{2} ||\L*\R^\top-\X_\star||_F^2,
    \label{equ:07}
\end{equation}
where $\X=\L*\R^\top$ with multi-rank $\r_m=[r,...,r]\in\mathbb{R}^{n_3}$. and $\L\in\mathbb{R}^{n_1\times r \times n_3},\R\in\mathbb{R}^{n_2\times r\times n_3}$.

First, it is easy to show that $f(X)$ satisfies restricted smoothness and restricted strong convexity.
\begin{lemma}
$f(\X):=\frac{1}{2} ||\X-\X_\star||_F^2$ satisfies the $(L,2r)$-restricted smoothness and $(\mu,2r)$-restricted strongly convexity with constants $\mu=1$ and $L=1$.
\label{lemma:08}
\end{lemma}

Therefore, based on the result of Lemma \ref{lemma:08} and Theorem \ref{theorem:main}, we establish linear convergence of APGD.
\begin{theorem}
Suppose that $||\X_0-\X_\star||_F\le \rho \sigma_{\min}(\bar{\bf{X}}_\star),\rho=\sqrt{\frac{1}{(1+4\kappa L/\mu)n_3}}$, and the step size $\eta\le 1 $, $\lambda_t\le  \sqrt{\frac{2}{c_2^2}}(f(\X_t)-f(\X_\star))^{1/2}$, then minimize (\ref{equ:07}) with APGD leads to 
\begin{align*}
    ||\X_t-\X_\star||_F^2 \le \left[1-\frac{\eta}{6}(1-\frac{\eta}{2}) \right]^{2t} ||\X_0-\X_\star||_F^2,
\end{align*}
where $c_2=(\frac{1}{\sqrt{5}-1} + \sqrt{\frac{8(s_{r}^m-s_{r_\star}^m)}{n_3}}).$
\end{theorem}
\begin{IEEEproof}
By combining the results of Theorem 2 and Lemma 2, we can directly obtain Theorem 3.
% As for the one-step convergence property, using Lemma 3 and Lemma 4, we have 
% $$
% f(\X_{t+1}) \le 
% $$
% It then follows that when $\eta = 1$, $f(X_t) \le 0$.
\end{IEEEproof}

\begin{remark}[\textbf{Comparison with exiting works}]
Previous works, such as Tubal-Alt-Min \cite{liu2019low} and ScaledGD \cite{wu2025guaranteed}, also achieved linear convergence in the exact-rank setting. However, none of them considered the over-parameterized case. For the first time, we establish a linear convergence rate under over-parameterization, significantly improving upon previous results.
\end{remark}

\subsection{Low-tubal-rank tensor recovery}
\label{sec:LTRTR}
The goal of low-tubal-rank tensor recovery is to recovery a tensor $\X_\star\in\mathbb{R}^{n_1\times n_2\times n_3}$ with tubal-rank $r_\star$ and multi-rank $\r_m^\star\in\mathbb{R}^{n_3}$ from a few linear measurements 
\begin{equation}
    \y=\M(\X_\star),
\end{equation}
where $\M(\cdot):\mathbb{R}^{n_1\times n_2\times n_3}\mapsto \mathbb{R}^m$ denotes the linear map. Based on the tensor BM factorization, we define the objective function:
\begin{equation}
f(\X) := \frac{1}{2}||\M(\X-\X_\star)||_2^2 =\frac{1}{2}\|\M(\L*\R^\top-\X_\star)\|_2^2,
\label{equ:06}
\end{equation}
In the low-tubal-rank tensor recovery problem, an important assumption is that the linear map satisfies the T-RIP condition, which is defined as follows.
\begin{definition}[Tensor Restricted
Isometry Property, T-RIP \cite{zhang2020rip}] A linear mapping, denoted as $\mathfrak{M}:\mathbb{R}^{n_1\times n_2 \times n_3}\to \mathbb{
R}^m$, is said to satisfy the T-RIP with parameter $(r,\ \delta_r)$ if
\begin{equation}
    (1-\delta_r)\| \X \|_F^2 \le \| \mathfrak{M}(\X) \|_2^2 \le (1+\delta_r)\| \X \|_F^2
\notag\end{equation}
holds for all tensors $\X\in \mathbb{R}^{n_1\times n_2 \times n_3}$ with a tubal-rank of at most $r$.
\end{definition}
 The T-RIP condition has been shown to hold with high probability if $m \gtrsim r(n_1 \lor n_2)n_3/\delta^2$, provided that each measurement tensor $\A_i$ in the operator $\M$ has entries drawn independently from a sub-Gaussian distribution with zero mean and variance $\frac{1}{m}$ \cite{zhang2021tensor} . Note that this condition has been extensively used in previous studies \cite{zhang2020rip,liu2024low,karnik2024implicit}, making it a natural and reasonable assumption in our setting.

If the linear map $\M$ satisfies the T-RIP condition, then $f(\X)$ fulfills the geometric assumptions stated in Theorem 2.
\begin{lemma}
Supposed that the linear map $\M$ satisfies the $(\delta,2r)$-T-RIP, then $f(\X)=\frac{1}{2}||\M(\X-\X_\star)||_2^2$ satisfies the $(L,2r)$-restricted smoothness and $(\mu,2r)$-restricted strongly convex with constants $\mu=1-\delta_{2r}$ and $L=1+\delta_{2r}$.
\label{lemma:04}
\end{lemma}

To satisfy the conditions of Theorem 2, we further need to ensure that the initialization method in Algorithm 2 meets the requirement $||\X_0 - \X_\star||_F \le \rho \sigma_{\min}(\bar{\bf{X}}_\star)$.
\begin{lemma}
Assuming that the linear map $\M$ satisfies $(\delta,2r)$-T-RIP with $\delta\le \frac{1}{\sqrt{8r\kappa^2n_3(1+12\kappa)}}$, taking the top tubal-rank $r$ truncated t-SVD of $\M^*(\y)$ as $\U_0*\S_0*\V_0^\top$, then the initialization \begin{equation}
    \L_0=\U_0*\S_0^{1/2},\ \R_0=\V_0*\S_0^{1/2}
    \label{equ:ini}
\end{equation} satisfies $\|\L_0*\R_0^\top - \X_\star\|_F \le \rho \sigma_{\min}(\bar{\bf{X}}_\star)$.
\label{lemma:05}
\end{lemma}
\begin{IEEEproof}
See Section VI of the supplementary material.
\end{IEEEproof}

Therefore, based on the result of Lemmas \ref{lemma:04}, \ref{lemma:05} and Theorem \ref{theorem:main}, we obtain the linear convergence for APGD.
\begin{theorem}
Suppose that the linear map $\M$ satisfies $(\delta,2r)$-T-RIP with $\delta\le \frac{1}{\sqrt{8r\kappa^2n_3(1+12\kappa)}}$, and the step size $\eta\le \frac{1}{1+\delta_{2r}}$, $\lambda_t\le  \sqrt{\frac{2}{(1+\delta)c_3^2}}(f(\X_t)-f(\X_\star))^{1/2}$, then minimizing $f(\X)$ with APGD and initialization (\ref{equ:ini}) leads to
\begin{align*}
   & f(\X_t) \le  (1-q_c)^{2t} f(\X_0) \\
    &||\X_t-\X_\star||_F^2 \le \frac{1+\delta}{1-\delta} (1-q_c)^{2t} ||\X_0-\X_\star||_F^2,
\end{align*}
where $c_3=\left(\frac{1}{\sqrt{5}-1}+ \frac{\sqrt{8(s_r^m-s_{r_\star}^m)}}{\sqrt{(1-\delta^2)n_3}}\right)$ and $q_c=(\eta -\frac{(1+\delta)\eta^2}{2})\frac{(1-\delta^2)}{6(1+\delta)}.$
\end{theorem}
\begin{IEEEproof}
Combining Lemmas 8 and 9, substituting $L=1+\delta$ and $\mu=1-\delta$ into Theorem 2 yields Theorem 4.

\end{IEEEproof}

\begin{remark}[\textbf{Comparison with existing works}]
The two most recent FGD-based methods for low-tubal-rank tensor recovery are \cite{liu2024low} and \cite{karnik2024implicit}. \cite{liu2024low} proves that FGD converges linearly in the exact-rank case but only sublinearly in the over-parameterized case; in both regimes, the rate depends on the condition number. \cite{karnik2024implicit} shows that with a very small initialization, FGD achieves linear convergence in both exact-rank and over-parameterized settings, but the required number of iterations scales as $\kappa^4$, which becomes prohibitive when the condition number is large. In contrast, our method attains linear convergence independent of the condition number and remains robust to over-parameterization. In addition, both of these works only consider the case where the target tensor is symmetric and positive semidefinite, whereas our method applies to arbitrary tensors.
\end{remark}

\section{Experiments}
\label{sec:exp}
In this section, we conduct simulation experiments to evaluate the performance of APGD on low-tubal-rank tensor decomposition and low-tubal-rank tensor recovery tasks. The results show that APGD achieves linear convergence even in over-parameterized and ill-conditioned cases, significantly outperforming other methods. In addition, we perform experiments to analyze several properties of APGD and discuss potential directions for future research.

\begin{figure*}[htbp]
\centering
\subfigure[]{
\begin{minipage}[t]{0.25\linewidth}
\centering
\includegraphics[width=4.5cm,height=4.5cm]{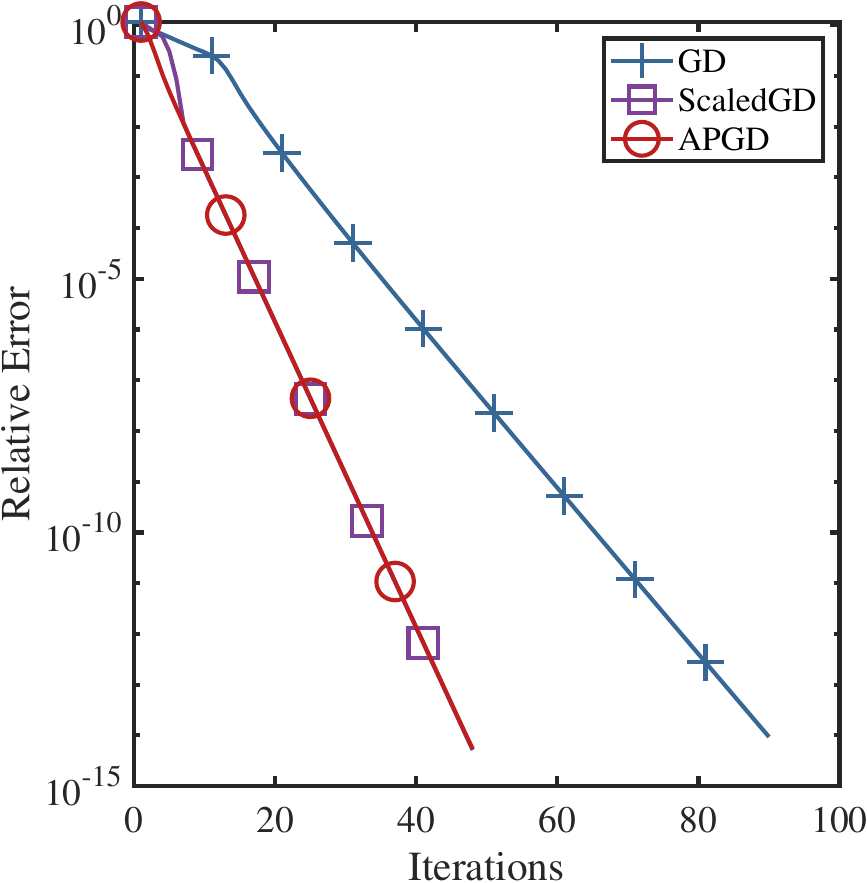}
\end{minipage}%
}%
\subfigure[]{
\begin{minipage}[t]{0.25\linewidth}
\centering
\includegraphics[width=4.5cm,height=4.5cm]{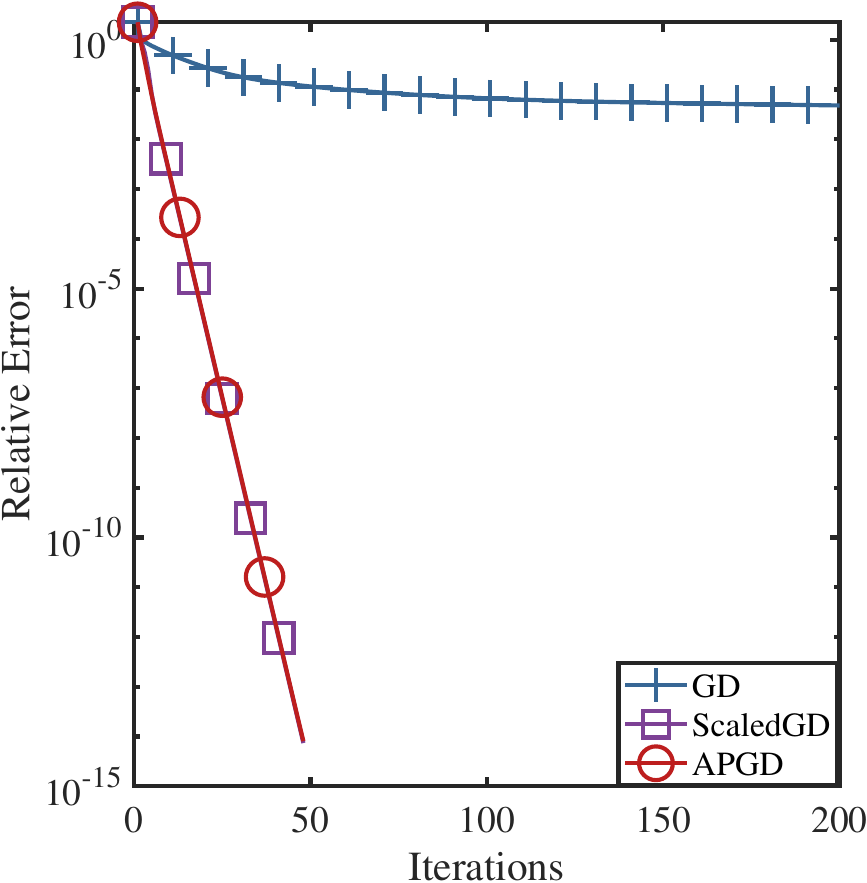}
\end{minipage}%
}%
\subfigure[]{
\begin{minipage}[t]{0.25\linewidth}
\centering
\includegraphics[width=4.5cm,height=4.5cm]{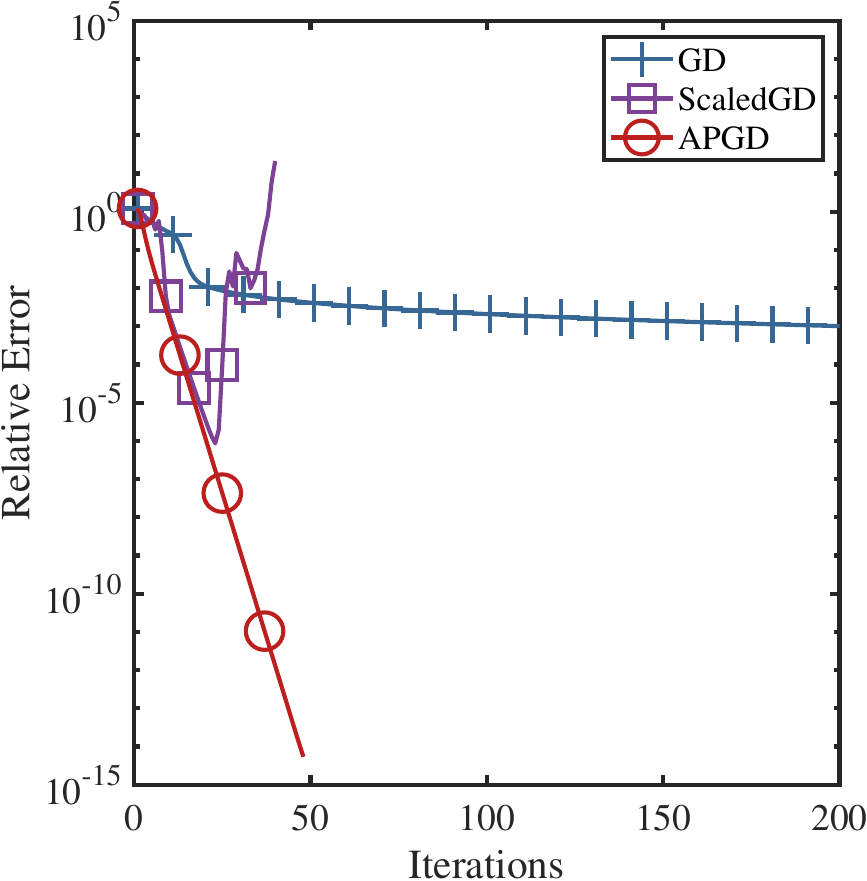}
\end{minipage}%
}%
\subfigure[]{
\begin{minipage}[t]{0.25\linewidth}
\centering
\includegraphics[width=4.5cm,height=4.5cm]{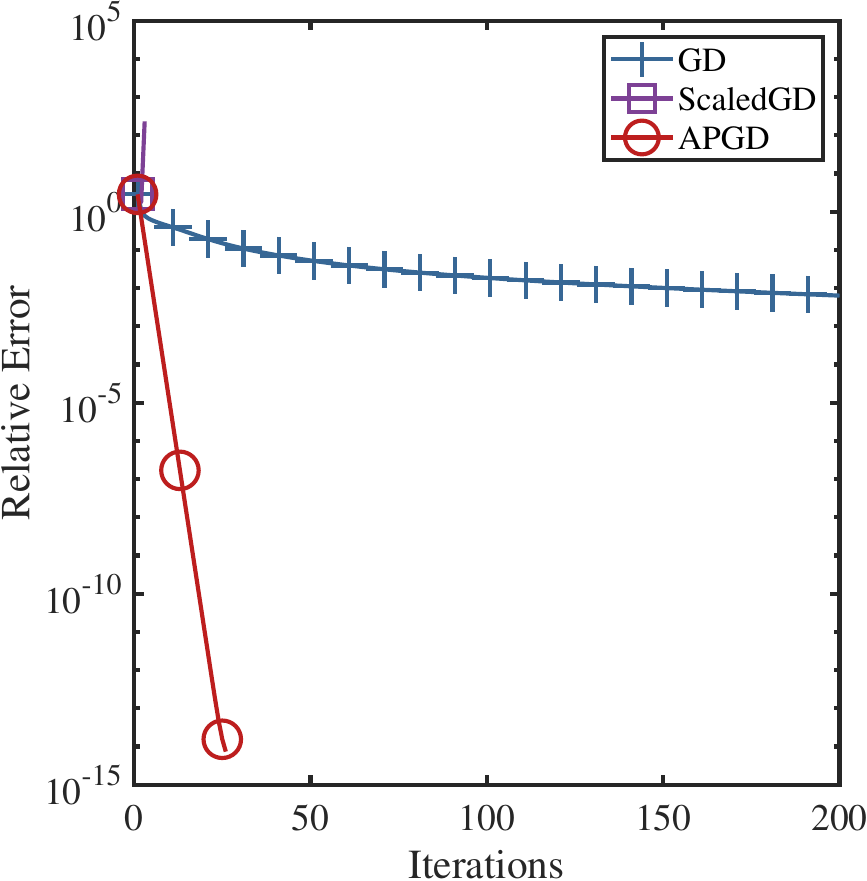}
\end{minipage}%
}%
\centering

\caption{  Relative recovery error of GD, ScaledGD, and APGD on the low-tubal-rank tensor factorization problem, where $n_1 = n_2 = 20,n_3=3$, $r_\star= 10$. The step size for all methods is set to 0.5.  In subfigure (a), the condition number of $\X_\star$ is 1, $r = r_\star$, and $\X_\star$ is full tubal-rank. In subfigure (b), the condition number of $\X_\star$ is 100, $r = r_\star$, and $\X_\star$ is full tubal-rank. In subfigure (c), the condition number of $\X_\star$ is 1, $r = r_\star$, but $\X_\star$ is not full tubal-rank. In subfigure (d), the condition number of $\X_\star$ is 1, $r = 2r_\star$, and $\X_\star$ is full tubal-rank. }
\label{fig:2}
\end{figure*}

\subsection{Simulations on low-tubal-rank tensor factorization}

First, we conduct simulation experiments on the low-tubal-rank tensor decomposition task. The ground-truth tensor $\X_\star$ is randomly generated with a tubal-rank of $r_\star$, and its condition number is controlled by adjusting the largest and smallest singular values. For all three methods, the initialization is random, i.e., $\L_0 \sim \mathcal{N}(0, 1/n_1)$ and $\R_0 \sim \mathcal{N}(0, 1/n_2)$. For APGD, we set $\lambda_t = f(X_t) / 10$.

From Figure \ref{fig:2}, we make the following observations:

1. GD is highly sensitive to the condition number, its convergence slows down as the condition number increases, whereas both ScaledGD and APGD are robust to it, demonstrating the effectiveness of preconditioning.

2. In both over-parameterized cases, ScaledGD diverges, while APGD still achieves linear convergence, confirming the robustness of APGD. Moreover, when the tubal-rank remains the same but $\X_\star$ is not of full tubal-rank, ScaledGD first converges and then diverges.

\begin{figure*}[htbp]
\centering
\subfigure[]{
\begin{minipage}[t]{0.25\linewidth}
\centering
\includegraphics[width=4.5cm,height=4.5cm]{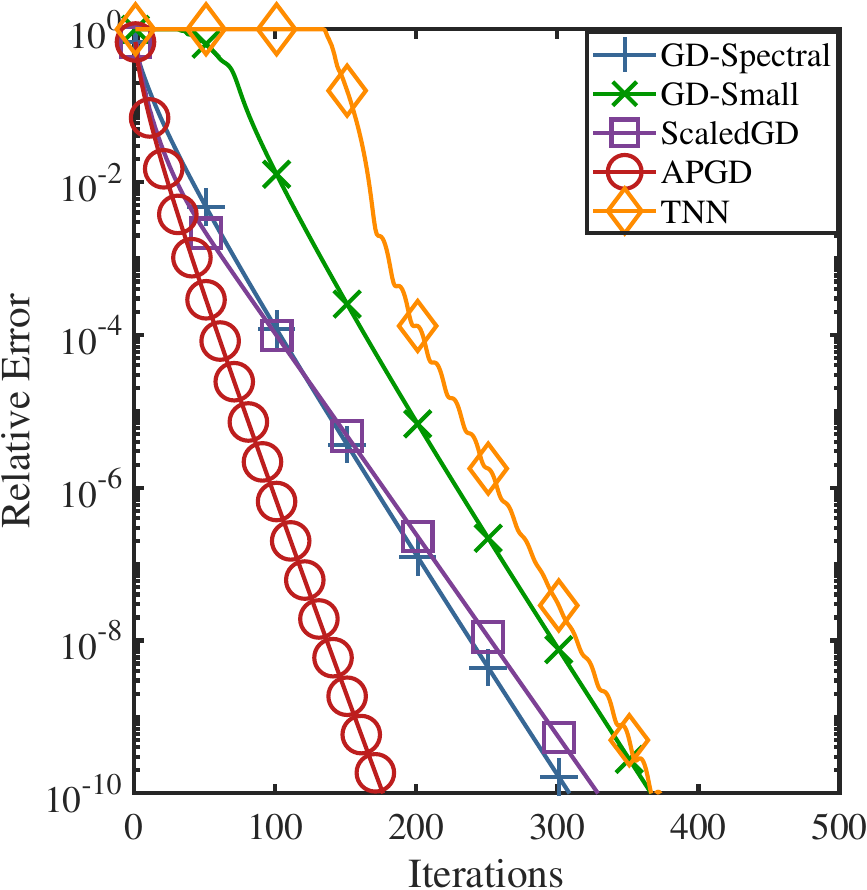}
\end{minipage}%
}%
\subfigure[]{
\begin{minipage}[t]{0.25\linewidth}
\centering
\includegraphics[width=4.5cm,height=4.5cm]{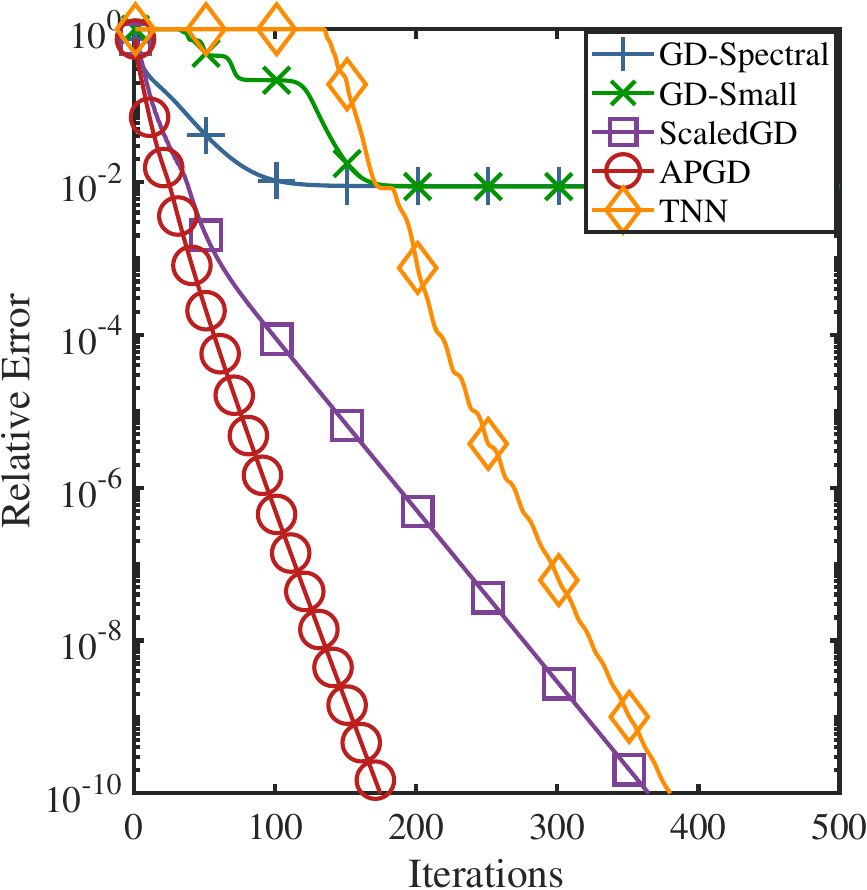}
\end{minipage}%
}%
\subfigure[]{
\begin{minipage}[t]{0.25\linewidth}
\centering
\includegraphics[width=4.5cm,height=4.5cm]{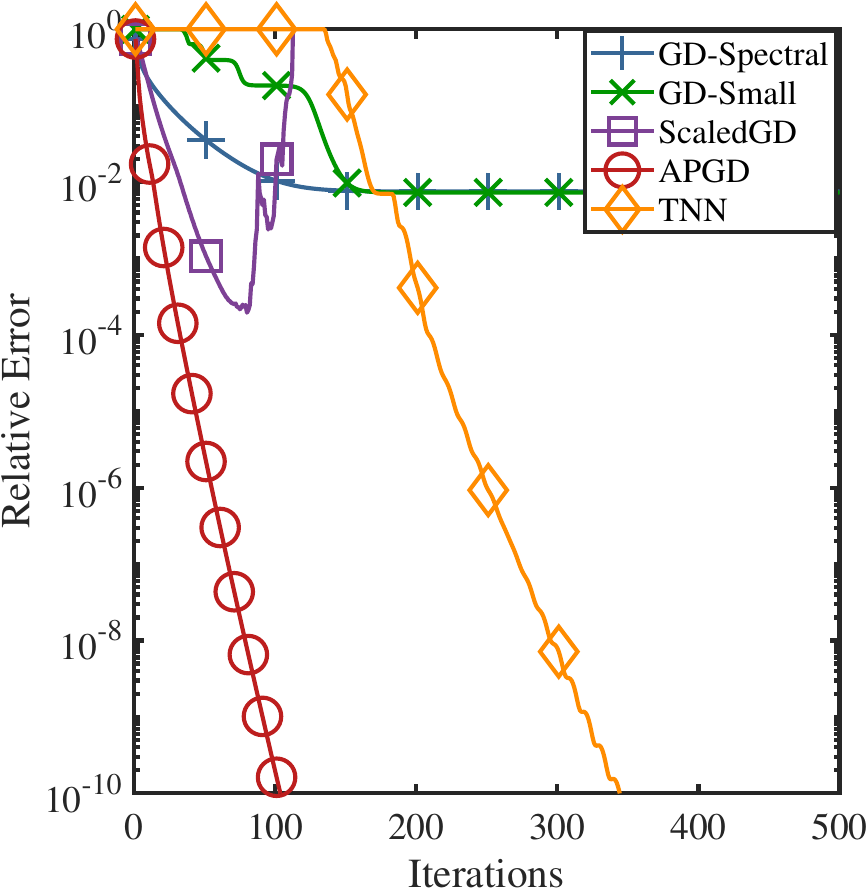}
\end{minipage}%
}%
\subfigure[]{
\begin{minipage}[t]{0.25\linewidth}
\centering
\includegraphics[width=4.5cm,height=4.5cm]{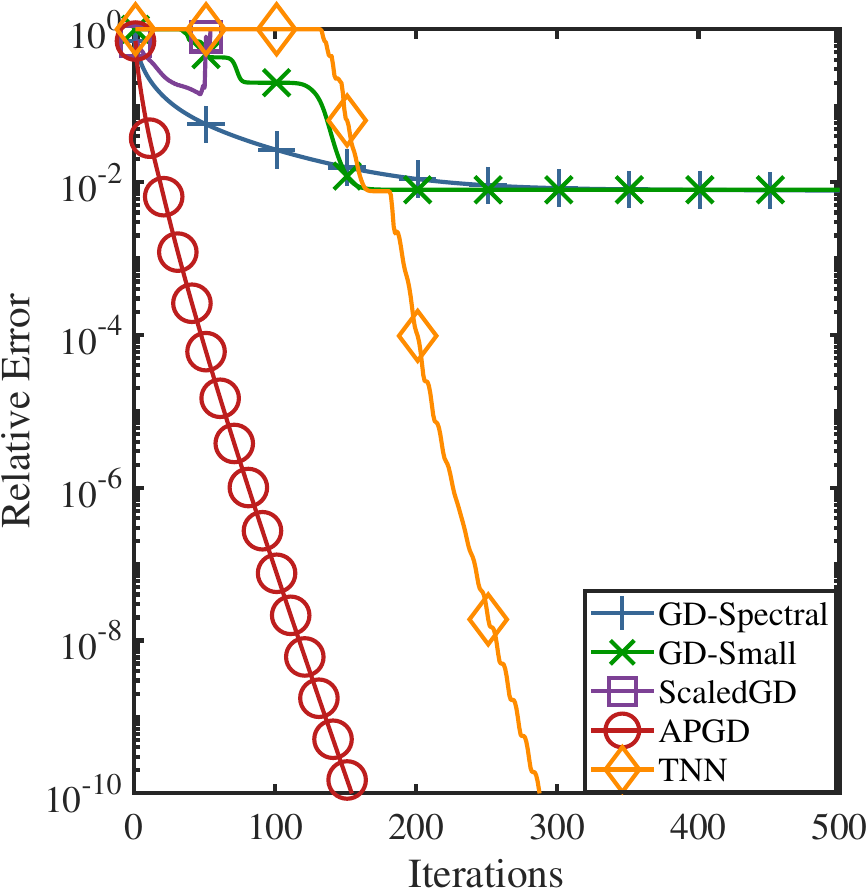}
\end{minipage}%
}%
\centering
 
\caption{Relative recovery error of different methods on the low-tubal-rank tensor recovery problem, where $n_1 = n_2 = 50,n_3=3$, $r_\star= 5$, $m=5rn_1n_3$. The step size of FGD-Spectral, FGD-Small, ScaledGD and APGD is set to 0.6. In subfigure (a), the condition number of $\X_\star$ is 2, $r = r_\star$, and $\X_\star$ is full tubal-rank. In subfigure (b), the condition number of $\X_\star$ is 100, $r = r_\star$, and $\X_\star$ is full tubal-rank. In subfigure (c), the condition number of $\X_\star$ is 100, $r = r_\star$, but $\X_\star$ is not full tubal-rank, its multi-rank $\r_m^\star=[1,5,5]$. In subfigure (d), the condition number of $\X_\star$ is 100, $r = 2r_\star$, and $\X_\star$ is full tubal-rank.}
\label{fig:3}
\end{figure*}

\begin{figure}[h]
\centering
\subfigure[]{
\begin{minipage}[t]{0.48\linewidth}
\centering
\includegraphics[width=4.3cm,height=4.3cm]{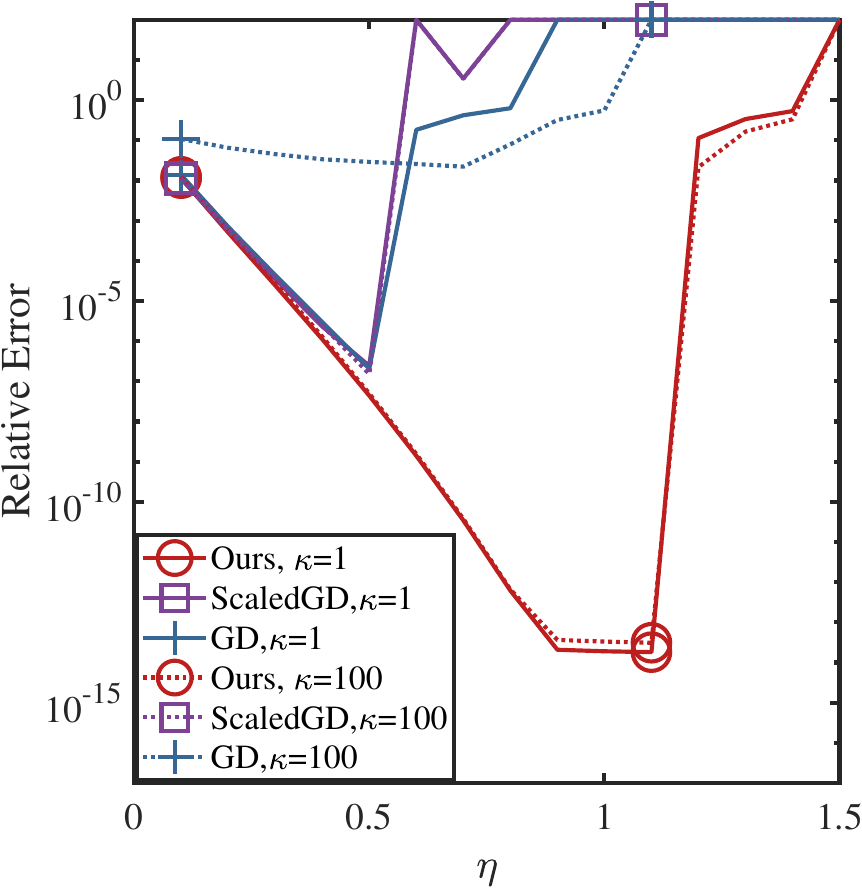}
\end{minipage}%
}
\subfigure[]{
\begin{minipage}[t]{0.48\linewidth}
\centering
\includegraphics[width=4.3cm,height=4.3cm]{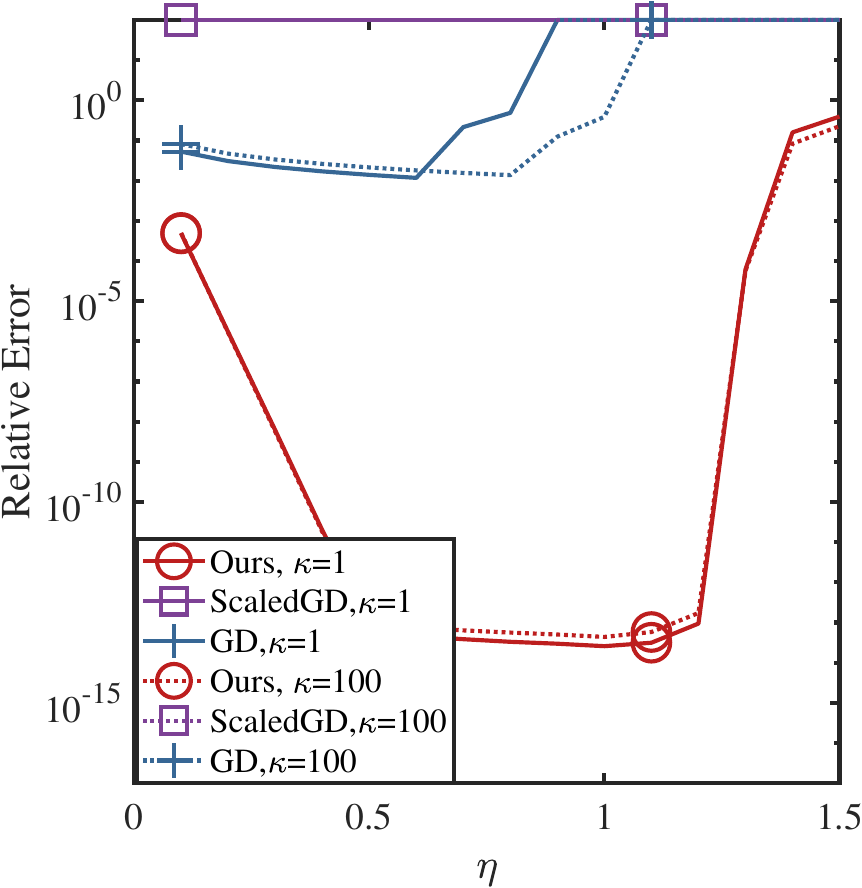}
\end{minipage}%
}
\centering
\caption{The relative error of APGD, ScaledGD, and GD after 100 iterations with respect to different step size $\eta$ under different condition numbers for low-tubal-rank tensor recovery. $n_1=n_2=20,,n_3=3, r_\star =10,\ m=5n_1n_3r$. Subfigure (a) denotes the exact rank case with $r=r_\star$ while subfigure (b) denotes the over-rank case with $r=2r_\star$.  }
\label{fig:4}
\end{figure}

\begin{figure}[h]
\centering
\subfigure[]{
\begin{minipage}[t]{0.48\linewidth}
\centering
\includegraphics[width=4.3cm,height=4.3cm]{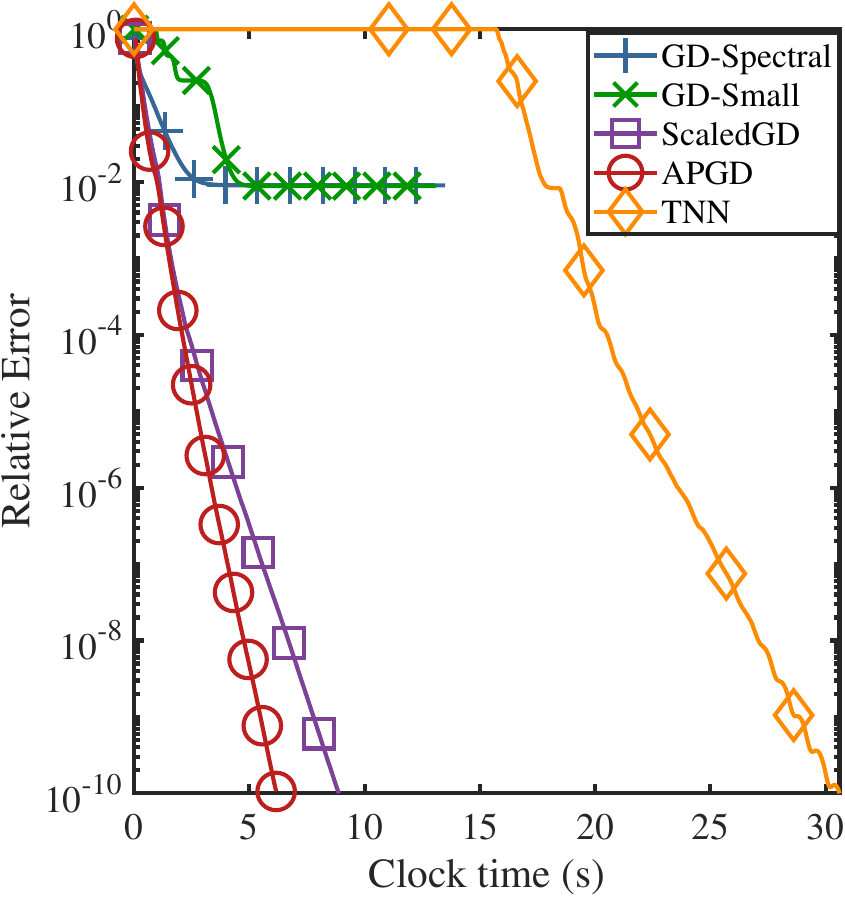}
\end{minipage}%
}
\subfigure[]{
\begin{minipage}[t]{0.48\linewidth}
\centering
\includegraphics[width=4.3cm,height=4.3cm]{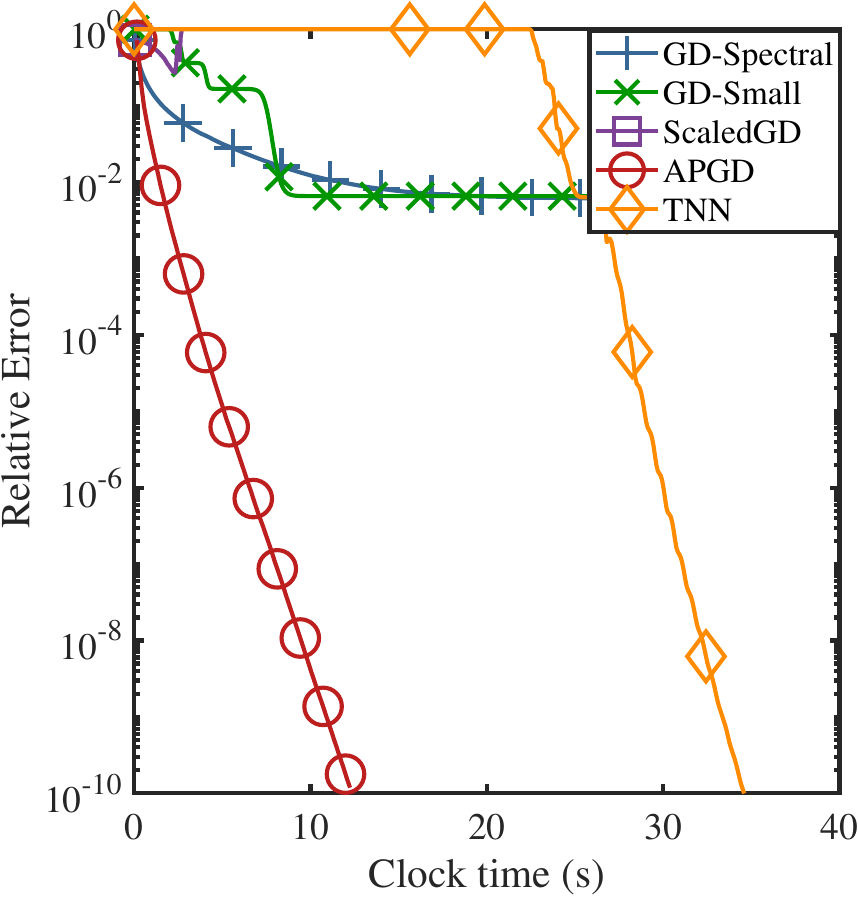}
\end{minipage}%
}
\centering
\caption{The running time of different methods  under different estimated tubal-rank $r$ for low-tubal-rank tensor recovery. $n_1=n_2=50,n_3=3, r_\star =5,\ m=5n_1n_3r$, $\eta=0.6$. Subfigure (a) denotes the exact rank case with $r=r_\star$ and $\r_\star^m=[5,5,5]$ while subfigure (b) denotes the over-rank case with $r=2r_\star$ and $\r_\star^m=[5,5,5]$.  }
\label{fig:5}
\end{figure}

\subsection{Simulations on low-tubal-rank tensor recovery}
We then evaluate the performance of APGD on the low-tubal-rank tensor recovery task, comparing with the classic convex method TNN \cite{lu2018exact}, FGD with spectral initialization \cite{liu2024low} and small initialization \cite{karnik2024implicit}. The ground-truth tensor $\X_\star$ is randomly generated, and each measurement tensor in the measurement operator $\M$ has entries sampled from a Gaussian distribution $\mathcal{N}(0, 1/m)$. The initialization for all three methods follows the spectral initialization described in Lemma \ref{lemma:05}. For APGD, we set $\lambda_t = f(\X_t) / 10$.

In Figure \ref{fig:3}, we plot the recovery error with different over-rank $r$ and condition number $\kappa$, and the following observations can be made:

1. In the exact-rank case, the convergence rate of GD is strongly affected by the condition number, while both ScaledGD and APGD are robust to it. Moreover, with the same step size, APGD converges slightly faster than ScaledGD.

2. When $r = r_\star$ but $\X_\star$ is not full tubal-rank, GD converges more slowly,  and ScaledGD starts to diverge once the error reaches around $10^{-7}$. This is caused by the increasing singularity of $(\L_t^\top * \L_t)$ and $(\R_t^\top * \R_t)$. In contrast, APGD remains stable due to the damping term.

3. When $r = 2r_\star$, the singularity of $(\L_t^\top * \L_t)$ and $(\R_t^\top * \R_t)$ becomes severe after only a few iterations, causing ScaledGD to diverge quickly. APGD, however, continues to converge linearly. In addition, since $r = 2r_\star$ increases the sample size, the convergence becomes faster. Consequently, the convergence rate of APGD in subfigure (d) is higher than that in subfigure (c).

4. The convex method TNN is robust across all settings, but it requires many more iterations to converge. Moreover, it shows almost no progress in the early stage and exhibits a sub-linear convergence rate.

\textbf{Verify the step size} In addition, we conduct experiments to compare the sensitivity of the three methods to the step size under different values of over-parameterized tubal-rank $r$ and condition number $\kappa$. As shown in Figure \ref{fig:4}, APGD exhibits the highest robustness to the step size. In both the exact-rank and over-parameterized cases, APGD remains convergent even when the step size exceeds 1. In the over-parameterized setting, the increased number of samples allows for a larger admissible step size than in the exact-rank case. Overall, APGD can use larger step sizes to achieve faster convergence. In contrast, ScaledGD is highly sensitive to the step size in both cases. In the exact-rank setting, the algorithm diverges when $\eta > 0.5$, and in the over-parameterized setting, it diverges even with a small step size of $\eta = 0.1$.
 GD is relatively less sensitive to the step size, but its convergence is too slow, 100 iterations are insufficient for GD to converge.

\textbf{Verify the computational time} 
Although APGD achieves the fastest convergence rate, its per-iteration complexity is higher than that of FGD and ScaledGD, which may lead to a larger total computational cost. However, APGD is more robust to the choice of step size and therefore allows the use of larger steps to accelerate convergence. Following Figure \ref{fig:4}, we selected for each method the largest step size that does not cause divergence, and we report the corresponding running times in Figure \ref{fig:5}. The results show that in the exact-rank setting, APGD requires less total computation time than ScaledGD, while FGD struggles to converge and TNN has the highest cost. In the over-parameterized setting, ScaledGD diverges, and APGD still achieves the smallest overall running time.

\subsection{Discussion of APGD}
In this section, we conduct simulation experiments on low-tubal-rank tensor recovery to analyze several properties of APGD. First, we verify the choice of the key parameter $\lambda$ and show that it is highly robust. Second, we analyze the dependence of APGD on initialization and show that APGD does not rely on spectral initialization but works well with arbitrary initializations. Finally,  we examine the dependence of APGD on the rebalancing process and demonstrate that rebalancing is unnecessary in practice. It is only required for theoretical proofs. 

\begin{figure}[h]
\centering
\subfigure[]{
\begin{minipage}[t]{0.48\linewidth}
\centering
\includegraphics[width=4.3cm,height=4.3cm]{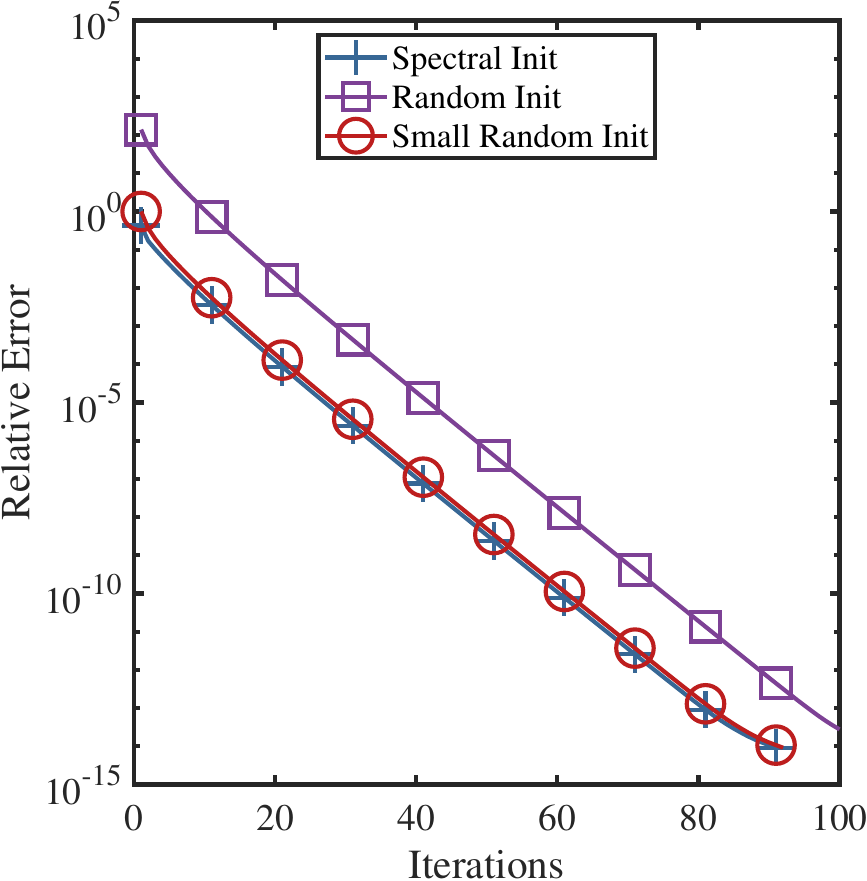}
\end{minipage}%
}
\subfigure[]{
\begin{minipage}[t]{0.48\linewidth}
\centering
\includegraphics[width=4.3cm,height=4.3cm]{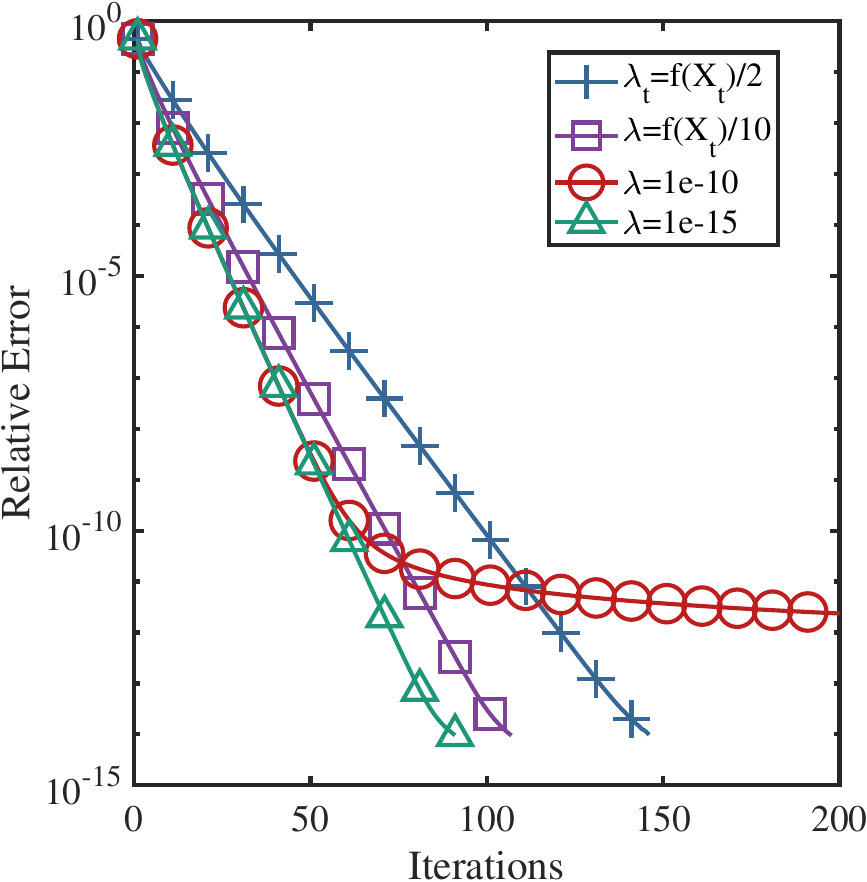}
\end{minipage}%
}
\centering
\caption{Recovery error of APGD under different initialization schemes and damping parameters $\lambda$. Here, $n_1 = n_2 = 20$, $n_3 = 3$, $r_\star = 10$, $r = 2r_\star$, $m = 5n_1n_3r$, and $\eta = 0.5$. Subfigure (a) shows the recovery error of APGD with different initialization methods, where “random init” denotes $\L_0 \sim \mathcal{N}(0,1)$ and $\R_0 \sim \mathcal{N}(0,1)$, and “small random init” denotes $\L_0 \sim \mathcal{N}(0,10^{-5})$ and $\R_0 \sim \mathcal{N}(0,10^{-5})$. Subfigure (b) shows the recovery error of APGD with different values of the damping parameter $\lambda$.}
\label{fig:6}
\end{figure}

\textbf{Verify initialization}
We first conduct experiments to examine the effect of different initialization methods on the convergence of APGD. Three initialization schemes are tested: spectral initialization, random initialization, and very small random initialization. Other experimental settings remain the same as in the previous subsection. As shown in subfigure (a) of Figure \ref{fig:6}, the convergence curves of APGD with small random initialization and spectral initialization almost overlap. When using random initialization, the initial error is larger, leading to more iterations, but the convergence rate remains the same. This demonstrates that APGD does not rely on spectral initialization and can still achieve linear convergence with random initialization. Therefore, extending the analysis from spectral to random initialization will be an important direction for future work.

\textbf{Verify damping parameter $\lambda$}
We then conduct experiments to examine the effect of the damping parameter $\lambda$. Four settings are tested: $\lambda_t = f(\X_t)/2$, $\lambda_t = f(\X_t)/10$, $\lambda = 10^{-10}$, and $\lambda = 10^{-15}$. Other experimental settings are the same as in the previous subsection. As shown in subfigure (b) of Figure \ref{fig:6}, the fastest convergence is achieved when $\lambda$ is set to a very small fixed value, followed by $\lambda_t = f(\X_t)/10$ and $\lambda_t = f(\X_t)/2$. When $\lambda = 10^{-10}$, APGD converges rapidly at first but slows down significantly once the error drops below $10^{-10}$. This indicates that $\lambda$ should be either extremely small or vary adaptively with the loss function, as discussed in Remark 3.3.

\begin{figure}[h]
\centering
\subfigure[]{
\begin{minipage}[t]{0.48\linewidth}
\centering
\includegraphics[width=4.3cm,height=4.3cm]{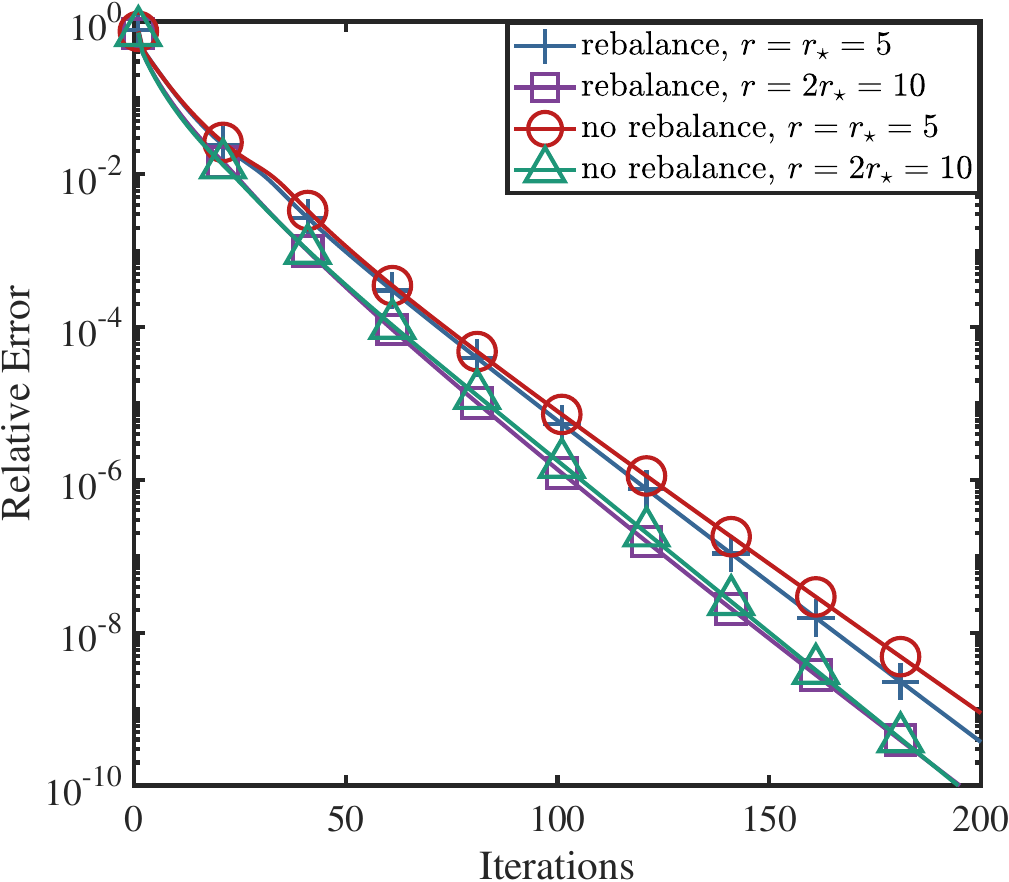}
\end{minipage}%
}
\subfigure[]{
\begin{minipage}[t]{0.48\linewidth}
\centering
\includegraphics[width=4.3cm,height=4.3cm]{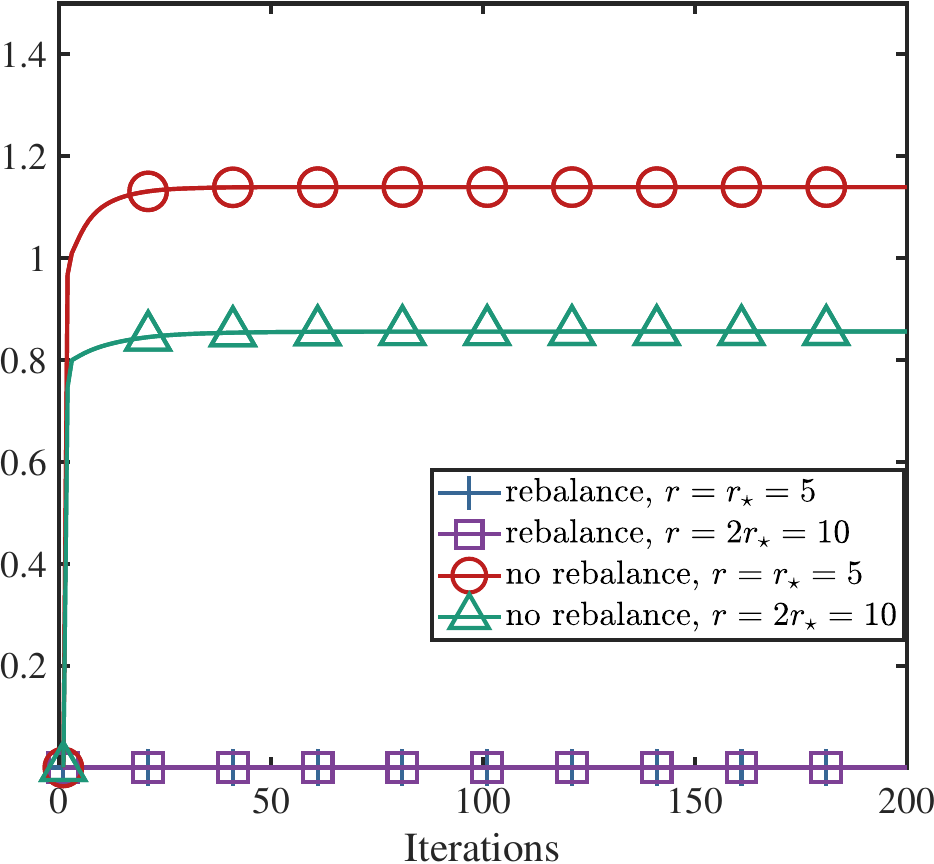}
\end{minipage}%
}
\centering
\caption{Comparison of APGD with and without the rebalancing step. Here, $n_1 = n_2 = 50$, $n_3 = 3$, $r_\star = 5$, $m = 5 n_1 n_3 r$, and $\eta = 0.5$. The initialization uses spectral initialization. Subfigure (a) shows the recovery error with and without the rebalancing step, and subfigure (b) reports $||\L^\top*\L-\R^\top*\R||_F$ in both cases.}
\label{fig:7}
\end{figure}

\textbf{Verify rebalancing process} Finally, we conduct experiments to evaluate the effect of the rebalancing procedure in the APGD algorithm. We compare the performance of APGD with and without rebalancing under both the exact-rank and over-parameterized settings.  
From Figure.\ref{fig:7}(a), we observe that adding or removing the rebalancing step has little effect on the convergence rate of APGD, indicating that the rebalancing step is optional from an algorithmic perspective. Figure.\ref{fig:7}(b) further shows that without the rebalancing step, the two factors remain  balanced. In particular, $||\L^\top * \L - \R^\top * \R||_F $ stays within a reasonably small bound. This suggests a promising direction for removing the rebalancing step theoretically by analyzing and controlling this mild imbalance directly. Some studies have explored how to remove the rebalancing step in the matrix sensing problem \cite{ma2021beyond,xiong2023over,soltanolkotabi2025implicit}, but research on this problem in the context of over-parameterized low-tubal-rank tensor estimation remains limited. Investigating how to remove the rebalancing step from the theoretical analysis will be an interesting direction for future work.

\section{Conclusion}
\label{sec:conclusion}
For the general low-tubal-rank tensor estimation problem, this paper proposes an Alternating Preconditioned Gradient Descent (APGD) algorithm that overcomes the limitations of previous tensor Burer-Monteiro-based methods such as Factorized Gradient Descent and ScaledGD in over-parameterized settings. We provide a rigorous convergence rate analysis to show that, under certain geometric assumptions on the objective function, APGD achieves linear convergence even when the model is over-parameterized. A series of simulations on low-tubal-rank tensor decomposition and low-tubal-rank tensor recovery tasks further verify the effectiveness of APGD. Finally, we discuss potential extensions of APGD, including the use of random initialization and the removal of the rebalancing step.

\bibliographystyle{IEEEtran}
\bibliography{reference}

\appendix
The Appendix is organized as follows:
\begin{itemize}
    \item Section \ref{sec:1} provides the  additional preliminaries;
    \item Section \ref{sec:2} provides the proof of  Lemma 2 in the main text;
    \item Section \ref{sec:3} provides the proof of  Lemma 4 in the main text;
    \item Section \ref{sec:4} provides the proof of  Lemma 5 in the main text;
    \item Section \ref{sec:5} provides the proof of  Lemma 6 in the main text;
    \item Section \ref{sec:6} provides the proof of  Lemma 9 in the main text.
\end{itemize}

\subsection{Preliminaries}
\label{sec:1}

Define tensor $\F_t=  \begin{bmatrix}
    \L_t \\ \R_t
\end{bmatrix} \in\mathbb{R}^{(n_1+  n_2)\times r \times n_3}$, $\L_t\in\mathbb{R}^{n_1\times r \times n_3}$, $\R_t \in\mathbb{R}^{n_2\times r \times n_3}$, $\X_t = \L_t*\R_t^\top$, where $\L_t$ and $\R_t$ denote the output of Algorithm 2. Note that we have $\L_t^\top*\L_t = \R_t^\top *\R_t$ due to the balance effect of Algorithm 2. And we define $\F_\star =\begin{bmatrix}
    \L_\star\\ \R_\star
\end{bmatrix}\in\mathbb{R}^{(n_1+  n_2)\times r_\star \times n_3}$, $\X_\star = \U_\star*\S_\star*\V_\star^\top$, $\L_\star=\U_\star*\S^{\frac{1}{2}}\in\mathbb{R}^{n_1\times r_\star \times n_3}$, $\R_\star=\V_\star*\S^{\frac{1}{2}}\in\mathbb{R}^{n_2\times r_\star \times n_3}$. For notation convenience, we define 
\begin{align*}
    \nabla f_t &= \nabla f(\X_t)=\nabla f(\L_t*\R_t^\top)\\
    \nabla_{\L} f_t &= \nabla_{\L} f(\X_t)=\nabla_{\L} f(\L_t*\R_t^\top)=\nabla f_t * \R_t\\
    \nabla_{\R} f_t &= \nabla_{\R} f(\X_t)=\nabla_{\R} f(\L_t*\R_t^\top) =\nabla f_t^\top *\L_t\\
    \P_L &:= \L^\top*\L+\lambda \I,\ \P_R:=\R^\top*\R+\lambda \I .
\end{align*}

Based on the above definitions, we present two important lemmas that will be used in our subsequent proofs.

\begin{lemma}
For $\F_t,\ \F_\star$ and $\X_t, \ \X_\star$, we have
\begin{equation}
||\F_t*\F_t^\top - \F_\star*\F_\star^\top ||_F \le 2 ||\X_t-\X_\star||_F.
\end{equation}
\label{lemma:1}
\end{lemma}
\begin{IEEEproof}
First we introduce two auxiliary tensors $\tilde{\F}_t:=\begin{bmatrix}
    \L_t\\ -\R_t
\end{bmatrix}$ and $\tilde{\F}_\star:=\begin{bmatrix}
    \L_\star\\ -\R_\star
\end{bmatrix}.$

Using the dilation trick, we have
\begin{equation}
\begin{aligned}
& \F_t*\F_t^\top - \tilde{\F}_t*\tilde{\F}_t^\top =2 \begin{bmatrix}
    0 & \X_t \\
    \X^\top_t & 0
\end{bmatrix}, \\
& \F_\star*\F_\star^\top - \tilde{\F}_\star*\tilde{\F}_\star^\top =2 \begin{bmatrix}
    0 & \X_\star \\
    \X^\top_\star & 0
\end{bmatrix}.
\end{aligned}
\end{equation}
Therefore, we have
\begin{equation}
\begin{aligned}
& 8|| \X_t-\X_\star ||_F^2\\
&\quad = || \F_t*\F_t^\top - \tilde{\F}_t*\tilde{\F}_t^\top - \F_\star*\F_\star^\top+ \tilde{\F}_\star * \tilde{\F}_\star^\top ||_F^2 \\
&\quad = || \F_t*\F_t^\top- \F_\star*\F_\star^\top ||_F^2 + || \tilde{\F}_t*\tilde{\F}_t^\top - \tilde{\F}_\star * \tilde{\F}_\star^\top||_F^2 \\
&\qquad - 2\langle \F_t*\F_t^\top- \F_\star*\F_\star^\top , \tilde{\F}_t*\tilde{\F}_t^\top - \tilde{\F}_\star * \tilde{\F}_\star^\top \rangle \\
&\quad \overset{(a)}{=} 2|| \F_t*\F_t^\top- \F_\star*\F_\star^\top ||_F^2 \\
&\qquad - 2\langle \F_t*\F_t^\top- \F_\star*\F_\star^\top , \tilde{\F}_t*\tilde{\F}_t^\top - \tilde{\F}_\star * \tilde{\F}_\star^\top \rangle \\
& \quad \overset{(b)}{=} 2|| \F_t*\F_t^\top- \F_\star*\F_\star^\top ||_F^2 + 2|| \F_t^\top * \tilde{\F}_\star ||_F^2 \\
&\qquad + 2|| \F^\top_\star * \tilde{\F}_t ||_F^2 \\
&\quad \ge 2|| \F_t*\F_t^\top- \F_\star*\F_\star^\top ||_F^2,
\end{aligned}
\label{equ:013}
\end{equation}
where (a) uses the fact that $$|| \F_t*\F_t^\top- \F_\star*\F_\star^\top ||_F^2 = || \tilde{\F}_t*\tilde{\F}_t^\top - \tilde{\F}_\star * \tilde{\F}_\star^\top||_F^2; $$ (b) use the fact that
$\F_t^\top *\tilde{\F}_t = \L^\top_t*\L_t -\R_t^\top*\R_t= 0 $ and $\F_\star^\top*\tilde{\F}_\star=\L_\star^\top *\L_\star -\R_\star^\top*\R_\star=0$. Therefore, we complete the proof of Lemma \ref{lemma:1}.
\end{IEEEproof}

\begin{lemma}
Assume that $f$ is $L$-smooth and $(\mu,\ 2r)$-restricted strongly convex, then we have
\begin{equation}
    \left|  \frac{2}{\mu + L} \langle \nabla^2 f(\X)[\Y],\Z \rangle - \langle \Y,\Z \rangle \right| \le \frac{L-\mu}{L+\mu} ||\Y||_F ||\Z||_F
\end{equation}
for all $\operatorname{rank}_t(\X)\le r$ and $\operatorname{rank}_t(\Y+\Z)\le 2r$, where $$\nabla^2 f(\X)[\Y]:\overset{\operatorname{def}}{=}\lim _{t\to 0} \frac{1}{t} [\nabla f(\X + t\Y)-\nabla f(\X)].$$
\label{lemma:09}
\end{lemma}
\begin{IEEEproof}
Assume that $f$ is $L$-smooth and $(\mu,\ 2r)$-restricted strongly convex, then we can derive that
\begin{equation}
\mu || \Y ||_F^2 \le \langle \nabla^2 f(\X)[\Y],\Y \rangle  \le L || \Y ||_F^2
\label{equ:04}
\end{equation}
for all $\operatorname{rank}_t(\X)\le r$ and $\operatorname{rank}_t(\Y)\le 2r$. 
Define $\H := \nabla ^2 f(\X)$, then one can verify that $\H$ is a linear operator and self-adjoint, i.e., $$\langle \H[\Y], \Z\rangle = \langle \Y, \H[\Z]\rangle$$and $$\H[\alpha_1 \Y_1 + \alpha_2 \Y_2] = \alpha_1 \H[\Y_1] +\alpha_2 \H[\Y_2].$$ Next, taking $\Y= \Y-\Z$ and $\Y = \Y+\Z$ into equation (\ref{equ:04}), we obtain
\begin{equation}
\mu || \Y+\Z ||_F^2 \le \underbrace{\langle \H[\Y+\Z],\Y+\Z \rangle }_{\C_1} \le L || \Y+\Z ||_F^2
\label{equ:7}
\end{equation}
and
\begin{equation}
\mu || \Y-\Z ||_F^2 \le \underbrace{\langle \H [\Y-\Z],\Y-\Z \rangle}_{\C_2}  \le L || \Y-\Z ||_F^2.
\label{equ:6}
\end{equation}

Without loss of generality, we can assume that $\|\Y\|_F = \|\Z\|_F = 1$, then we have 
$|| \Y+\Z ||_F^2 = 2+ 2 \langle \Y, \Z \rangle,\ || \Y-\Z ||_F^2 = 2 - 2 \langle \Y, \Z \rangle.$ Subtracting Equation (\ref{equ:6}) from Equation (\ref{equ:7}), we obtain
\begin{equation}
\C_1-\C_2 = 4\langle \H[\Y],\Z \rangle.
\end{equation}
Then we have 
\begin{equation}
\begin{aligned}
\C_1-\C_2&\ge \mu || \Y+\Z ||_F^2 - L || \Y-\Z ||_F^2\\
&=\mu (2+ 2 \langle \Y, \Z \rangle) - L (2 - 2 \langle \Y, \Z \rangle)  \\
&=2(\mu-L) + 2(\mu+L)\langle \Y, \Z \rangle\\
\C_1-\C_2 &\le  L || \Y+\Z ||_F^2 - \mu || \Y-\Z ||_F^2 \\
&= L(2+2\langle \Y, \Z \rangle) - \mu (2-2\langle \Y, \Z \rangle) \\
&=2(L-\mu) + 2(L+\mu)\langle \Y, \Z \rangle,
\end{aligned}
\end{equation}
which leads to
\begin{equation}
\frac{(\mu-L)}{2} \le \langle \H[\Y],\Z \rangle -  \frac{(\mu+L)}{2}\langle \Y, \Z \rangle \le  \frac{(L-\mu)}{2}.
\label{equ:010}
\end{equation}
Multiplying Equation (\ref{equ:010}) by $\frac{2}{\mu+L}$, we obtain:
\begin{equation}
- \frac{L-\mu}{L+\mu} \le \frac{2}{(\mu+L)}  \langle \H[\Y],\Z \rangle -  \langle \Y, \Z \rangle \le \frac{L-\mu}{L+\mu},
\end{equation}
which is exactly 
$$
    \left|  \frac{2}{\mu + L} \langle \nabla^2 f(\X)[\Y],\Z \rangle - \langle \Y,\Z \rangle \right| \le \frac{L-\mu}{L+\mu} ||\Y||_F ||\Z||_F
$$ under the assumption $||\Y||_F = ||\Z||_F = 1$. Therefore, we complete the proof of Lemma \ref{lemma:09}. 
\end{IEEEproof}

\subsection{Proof of Lemma 2}
\label{sec:2}
\begin{IEEEproof}
Note that 
\begin{align*}
&(\L+ \D_1)*(\R+ \D_2)^\top-\X_\star \\
&= \underbrace{\L*\R^\top -\X_\star}_{\K_1}+ \underbrace{\D_1*\R^\top + \L*\D_2^\top}_{\K_2} +  \underbrace{\D_1*\D_2^\top}_{\K_3}.
\end{align*}
Then \begin{align*}
g(\Z+\D) &= \frac{1}{2}||\K_1+\K_2+\K_3||_F^2\\
&=\frac{1}{2}\left[ ||\K_1||_F^2 + ||\K_2||_F^2+ ||\K_3||_F^2 \right] \\
&+ \langle \K_1, \K_2\rangle +\langle \K_1, \K_3\rangle + \langle \K_2, \K_3\rangle,
\end{align*}
where $\langle \K_1, \K_2\rangle=\langle \nabla g(\Z), \D \rangle.$

For $||\K_2||_F^2$, we have
\begin{align*}
&||\K_2||_F^2 =||\D_1*\R^\top + \L*\D_2^\top||_F^2 \\
&=||\D_1*\R^\top||_F^2+||\L*\D_2^\top||_F^2+ 2\langle\D_1*\R^\top ,\L*\D_2^\top\rangle\\
&\le 2 ||\D_1*\R^\top||_F^2 + 2 ||\L*\D_2^\top||_F^2\\
&=2 ||\D_1*(\R^\top*\R)^{\frac{1}{2}} * (\R^\top*\R)^{-\frac{1}{2}} *\R^\top||_F^2\\
&+2||\L*(\L^\top*\L)^{-\frac{1}{2}}* (\L^\top*\L)^{\frac{1}{2}}*\D_2^\top ||_F^2 \\
&\le 2 ||\D_1*(\R^\top*\R)^{\frac{1}{2}} ||_F^2 || (\R^\top*\R)^{-\frac{1}{2}}*\R^\top||_2^2\\
&+ 2||(\L^\top*\L)^{\frac{1}{2}}*\D_2^\top||_F^2 ||\L*(\L^\top*\L)^{-\frac{1}{2}}||_2^2\\
&\overset{(1)}{\le} 2||\D||_P^2,
\end{align*}
where (1) use the fact that $||\L*(\L^\top*\L)^{-\frac{1}{2}}||_2=|| (\R^\top*\R)^{-\frac{1}{2}}*\R^\top||_2=1$.

For $|| \K_3||_F$, we have
\begin{align*}
&||\K_3||_F^2 \le ||\D_1||^2_P ||\D_2||^2_P ||(\L^\top * \L)^{-\frac{1}{2}}||_2^2 ||(\R^\top * \R)^{-\frac{1}{2}}||_2^2 \\
&= ||\D_1||^2_P ||\D_2||^2_P/(\sigma^2_r(\bar{\bf{L}}) * \sigma^2_r(\bar{\bf{R}}))\\
&\le ||\D||_P^4 / (2\sigma^4_{\min}). 
\end{align*}

For $\langle \K_1, \K_3\rangle$, we have
\begin{align*}
& \langle \K_1, \K_3\rangle\le ||\L*\R^\top -\X_\star||_F ||\D_1*\D_2^\top||_F\\
& \le ||\L*\R^\top -\X_\star||_F  ||\D||_P^2 / (\sqrt{2}\sigma^2_{\min}).
\end{align*}
For $\langle \K_2, \K_3\rangle$, we have
\begin{align*}
&\langle \K_2, \K_3\rangle \le ||\D_1*\R^\top + \L*\D_2^\top||_F || \D_1*\D_2^\top||_F\\
&\le ||\D||_P^3/\sigma_{\min}^2 . 
\end{align*}

Combining all these terms, we obtain 
\begin{align*}
    g(\Z+\D) = g(\Z) + \langle \nabla g(\Z) , \D\rangle +  \frac{L_P}{2}||\D||_P^2,
\end{align*}
where $L_P=2+ \frac{||\D||_P^2}{2\sigma^4_{\min}} + \frac{\sqrt{2}||\E||_F}{\sigma^2_{\min}} + \frac{2||\D||_P}{\sigma^2_{\min}}.$

\end{IEEEproof}

\subsection{Proof of Lemma 4}
\label{sec:3}
We first present a variational form of the gradient norms to facilitate the subsequent derivation of a PL-type inequality.
\begin{lemma}
Assume that $f$ is $L$-smooth and $(\mu,\ 2r)$-restricted strongly convex, then we have
\begin{align*}
&||\nabla_{\L}f(\X)||_F\\
&\qquad \ge \upsilon \underset{||\O_1||_F=1}{\max}[\langle \E, \O_1*\R^\top \rangle-\varepsilon ||\E||_F||\O_1*\R^\top||_F ]\\
&||\nabla_{\R}f(\X)||_F\\
&\qquad \ge \upsilon \underset{||\O_2||_F=1}{\max}[\langle \E, \L*\O_2^\top \rangle-\varepsilon ||\E||_F||\L*\O_2^\top||_F ],
\end{align*}
where $\upsilon=\frac{L+\mu}{2}$, $\varepsilon=\frac{L-\mu}{L+\mu}$ and $\E=\X-\X_\star$.
\label{lemma:010}
\end{lemma}
\begin{IEEEproof}
Define $G(\X):=\nabla f(\X)$, then for any $\X\in\mathbb{R}^{n_1\times n_2\times n_3}$, we have
\begin{equation}
\begin{aligned}
 G(\X)-G(\X_\star) = \int_0^1 \nabla ^2 f(\X_\star + t\E)[\E] dt,\ \E=\X-\X_\star.
\end{aligned}
\end{equation}
Therefore, for any $||\O_1||_F=1$,
\begin{align*}
\langle \nabla_{\L} f(\X), \O_1 \rangle &= \langle G(\X)-G(\X_\star) ,\O_1*\R^\top \rangle\\
& = \int_0^1  \langle  \nabla ^2 f(\X_\star +t\E)[\E], \O_1*\R^\top \rangle dt\\
&\overset{(1)}{\ge} \upsilon [\langle \E, \O_1*\R^\top \rangle-\varepsilon ||\E||_F||\O_1*\R^\top||_F ],
\end{align*}
where (1) uses the result of Lemma \ref{lemma:09} and the fact that $\operatorname{rank}_t(\X_\star+t\E)\le 2r$.
Note that 
\begin{align*}
||\nabla_{\L}f(\X)||_F=\underset{||\O_1||_F=1}{\max} \langle \nabla_{\L}f(\X), \O_1\rangle,
\end{align*}
therefore we prove the first inequality in Lemma \ref{lemma:010} and the second inequality can be proved by an analogous argument.

\end{IEEEproof}

Building on Lemma 12, we now prove Lemma 4.
\begin{lemma}[Rewrite Lemma 4]
Assume that $f$ is $L$-smooth and $(\mu,\ 2r)$-restricted strongly convex, then we have
\begin{align*}
&\frac{||\nabla_{\L}f(\L*\R_t^\top)||_{P_R^*}}{||\L*\R^\top-\X_\star||_F}\ge \underset{k\in\{1,..,s_r^m\}}{\max} \frac{\mu+L}{2}    \frac{\cos\vartheta_{\R}^k-\varepsilon}{\sqrt{1+\lambda/\sigma_{k}^2(\bar{\bf{R}})}}\\
&\frac{||\nabla_{\R}f(\L*\R_t^\top)||_{P_L^*}}{||\L*\R^\top-\X_\star||_F}\ge \underset{k\in\{1,..,s_r^m\}}{\max} \frac{\mu+L}{2}  \frac{\cos\vartheta_{\L}^k-\varepsilon}{\sqrt{1+\lambda/\sigma_{k}^2(\bar{\bf{L}})}},
\end{align*}
where $\varepsilon=\frac{L-\mu}{L+\mu}$ and $\cos\vartheta_{\R}^k,\ \cos\vartheta_{\L}^k$ are defined as
\begin{align*}
&\cos\vartheta_{\R}^k= \max_{\O_1\in\mathbb{R}^{n_1\times r\times n_3}} \frac{\langle \bf{\bar{E}},\bf{\bar{O}_1} \bar{\bf{R}}_k\rangle}{||\bar{\bf{E}}||_F ||\bar{\bf{O}}_1\bar{\bf{R}}^\top_k||_F},\\
&\cos\vartheta_{\L}^k= \max_{\O_2\in\mathbb{R}^{n_2\times r\times n_3}} \frac{\langle \bar{\bf{E}},\bar{\bf{L}}\bar{\bf{O}}_2^\top \rangle}{||\bar{\bf{E}}||_F ||\bar{\bf{O}}_2^\top\bar{\bf{L}}_k||_F},
\end{align*}
and $\bar{\bf{R}}_k$ and $\bar{\bf{L}}_k$ denote the top-$k$ svd truncation of $\bar{\bf{R}}$ and  $\bar{\bf{L}}$, and $s_r^m=||\r_m||_1$.
\end{lemma}

\begin{IEEEproof}
Repeat the proof of Lemma 12, we have
\begin{equation}
\begin{aligned}
 &||\nabla_{\L}f(\L*\R_t^\top)||_{P_R^*} \\
&\ge \upsilon \left\{ \underset{||\O||_{P_R}=1}{\max} \langle \E , \O_1*\R^\top \rangle -\varepsilon ||\E||_F||\O_1*\R^\top||_F \right\}\\
&= \frac{\upsilon}{n_3} \left\{ \underset{||\O||_{P_R}=1}{\max} \langle \bar{\bf{E}} , \bar{\bf{O}}_1 \bar{\bf{R}}^\top \rangle -\varepsilon ||\bar{\bf{E}}||_F||\bar{\bf{O}}_1 \bar{\bf{R}}^\top||_F \right\},
\end{aligned}
\end{equation}
where $\upsilon=\frac{L+\mu}{2}$.
For any $k\in \{1,2,...,s_r^m\}$, we can restrict this problem so that
\begin{align*}
 &||\nabla_{\L}f(\L*\R_t^\top)||_{P_R^*}\\
 &\ge \frac{\upsilon}{n_3} \left\{ \underset{||\O||_{P_R}=1}{\max} \langle \bar{\bf{E}} , \bar{\bf{O}}_1 \bar{\bf{R}}_k^\top \rangle -\varepsilon ||\bar{\bf{E}}||_F||\bar{\bf{O}}_1 \bar{\bf{R}}^\top_k||_F \right\}\\
 &\ge \frac{\upsilon}{n_3} ||\bar{\bf{E}}||_F||\bar{\bf{O}}_1 \bar{\bf{R}}^\top_k||_F (\cos\vartheta_{R}^k-\varepsilon)\\
 &= \upsilon ||\E||_F||\O_1^{\star}*\mathfrak{J}_k(\R)^\top||_F (\cos\vartheta_{R}^k-\varepsilon),
\end{align*}
where $ \mathtt{bdiag}(\overline{\mathfrak{J}_k(\R)})=\bar{\bf{R}}_k $ and $\O_1^\star$ denotes the solution of $\cos\vartheta _{\R}^k$.
Note that 
\begin{equation}
\begin{aligned}
 ||\O_1^{\star}*\mathfrak{J}_k(\R)^\top ||^2_F & = ||\O_1^{\star}*\P_{\R}^{\frac{1}{2}}*\P_{\R}^{-\frac{1}{2}} *\mathfrak{J}_k(\R)^\top||^2_F\\
 & \overset{(1)}{\ge} \sigma^2_{\min}(\overline{P_{R}}^{-\frac{1}{2}} \cdot \overline{\mathfrak{J}_k(R)}^\top) ||\O_1^{\star}*\P_{\R}^{\frac{1}{2}}||^2_F \\
 &\overset{(2)}{\ge} \sigma^2_{\min}(\overline{P_{R}}^{-\frac{1}{2}} \cdot \overline{\mathfrak{J}_k(R)}^\top)\\
 &\overset{(3)}{\ge} \frac{1}{1+\lambda/\sigma^2_k(\bar{\bf{R}})},
\end{aligned}
\end{equation}
where (1) uses the fact that $||\A*\B||_F\ge\sigma_{\min}(\bar{\bf{A}})||\B||_F$; (2) uses the fact that $||\O_1^{\star}*\P_{\R}^{\frac{1}{2}}||^2_F = ||\O_1||_{P_R}=1$; (3) use that fact that the singular value of $\frac{\sigma_k(\bar{R})}{\sqrt{\sigma^2_k(\bar{R})}+\lambda}$.

Substituting these together yields
$$
\frac{||\nabla_{\L}f(\L*\R_t^\top)||_{P_R^*}}{||\L*\R^\top-\X_\star||_F}\ge \underset{k\in\{1,..,s_r^m\}}{\max} \frac{\mu+L}{2}    \frac{\cos\vartheta_{\R}^k-\varepsilon}{\sqrt{1+\lambda/\sigma_{k}^2(\bar{\bf{R}})}}.
$$
Another inequality can be proved by an analogous argument.
\end{IEEEproof}

\subsection{Proof of Lemma 5}
\label{sec:4}
To prove Lemma 5, we first construct a symmetric tensor and then derive the corresponding upper bound on $\sin\theta$.
\begin{lemma}
For $\bm{\mathcal{M}}_\star = \G * \G ^\top \in\mathbb{R}^{n \times n \times n_3}$, one supposes that $\F\in\mathbb{R}^{n \times r \times n_3}$ satisfies $||\F * \F^\top - \bm{\mathcal{M}}_\star  ||_F \le \rho \sigma_{\min} (\bm{\bar{M}}_\star)$ with $\rho= \frac{1}{\sqrt{2n_3}}$ and $r_\star = \operatorname{rank}_t(\bm{\mathcal{M}}_\star)$. Then we have
\begin{equation}
\begin{aligned}
\sin\theta &  = \frac{|| (\I-\F*\F^{\dagger}) * \bm{\mathcal{M}}_\star *  (\I - \F * \F^{\dagger}) ||_F}{|| \F*\F^\top - \bm{\mathcal{M}}_\star ||_F} \\
& \le \frac{1}{\sqrt{2}} \frac{\rho}{\sqrt{1-\rho^2}}.\\
\end{aligned}
\notag
\end{equation}
\label{lemma:23}
\end{lemma}
\begin{IEEEproof}
For $\F\in\mathbb{R}^{n\times r \times n_3}$ and $\G\in\mathbb{R}^{n\times r_\star \times n_3}$, suppose that $\F$ satisfies
\begin{equation}
\rho \overset{\operatorname{def} }{ = } \frac{ || \F*\F ^\top - \bm{\mathcal{M}}||_F  }{\sigma_{\min}(\bf{\bar{M}_\star})} < \frac{1}{\sqrt{2n_3}}. 
\label{equ:57}
\end{equation}
Then we have
\begin{align*}
\sigma_{s_{r^\star}^m} (\bar{\bf{F}}\bar{\bf{F}}^\top)&=\sigma_{s_{r^\star}^m} (\bar{\bf{M}}+\bar{\bf{F}}\bar{\bf{F}}^\top-\bar{\bf{M}}) \\
&\ge \sigma_{s_{r^\star}^m}(\bar{\bf{M}})-||\bar{\bf{F}}\bar{\bf{F}}^\top-\bar{\bf{M}}||_F\\
& = \sigma_{s_{r^\star}^m}(\bar{\bf{M}})-\sqrt{n_3}|| \F*\F ^\top - \bm{\mathcal{M}}||_F\\
&\ge (1-\frac{1}{\sqrt{2}}) \sigma_{s_{r^\star}^m}(\bar{\bf{M}}).
\end{align*}

Next, leveraging the rotational invariance of the problem, we can, without loss of generality, assume that 
\begin{equation}
\begin{aligned}
\F = \begin{bmatrix}
    \F_1&0\\
    0& \F_2
\end{bmatrix},\ \G =\begin{bmatrix}
    \G_1\\
    \G_2
\end{bmatrix},\ \sigma_{\min}(\bar{\bf{F}}_1)\ge \sigma_{\max}(\bar{\bf{F}}_2)
\end{aligned}
\label{equ:10}
\end{equation}
where $\F_1\in\mathbb{R}^{r \times r \times n_3},\ \F_2 \in\mathbb{R}^{(n-r)\times r \times n_3}$, $\G_1 \in\mathbb{R}^{r\times r_\star \times n_3}, \G_2\in\mathbb{R}^{(n-r)\times r_\star \times n_3}$. Then the $\sin\theta $ satisfies
\begin{equation}
\begin{aligned}
&||\E||_F \sin\theta \\
&\ \ = \underset{\Y}{\min} ||(\F*\Y^\top+\Y*\F^\top)-(\F*\F^\top-\G*\G^\top)||_F \\
&\ \ = \underset{\Y_1,\Y_2}{ \min} \left \|  \Z_1 -  \Z_2\right\|_F \\
&=||{\G_2} * {\G_2}^\top ||_F = || (\I - \F*\F^{\dagger})*\G*\G*(\I-\F*\F^{\dagger})  ||_F,
\end{aligned}
\end{equation}
where
\begin{equation}
\begin{aligned}
&\Z_1= \begin{bmatrix}
    \F_1*\Y_1^\top + \Y_1*\F_1^\top &  \F_1*\Y_2^\top\\
    \Y_2*\F_1^\top & 0
\end{bmatrix} \\
&\Z_2 = \begin{bmatrix}
    \F_1*\F_1^\top-\G_1 *{\G_1}^\top& \G_2*{\G_1}^\top\\
    \G_1*{\G_2}^\top& -\G_2*{\G_2}^\top
\end{bmatrix}.\\
\end{aligned}
\notag
\end{equation}
Before proving the upper bound of $\sin\theta$, we first present and prove a lemma that will be useful later.
\begin{lemma}
Suppose that $\F,\ \G$ are defined as (\ref{equ:10}) and $\rho$ satisfies the assumption in Lemma \ref{lemma:23}, then we have $\lambda_{\min}({\bar{\bf{G}}_1}^\top\bar{\bf{G}}_1
) \ge \lambda_{\max} ({\bar{\bf{G}}_2}^\top\bar{\bf{G}}_2)$.
\label{lemma:9}
\end{lemma}
\begin{IEEEproof}
Define $\gamma_1 = \lambda_{\min}(\bar{\bf{G}}_1^\top \bar{\bf{G}}_1)$ and $\gamma_2 = \lambda_{\max}(\bar{\bf{G}}_2^\top \bar{\bf{G}}_2)$. We are going to prove that $\gamma_1 \ge \gamma_2$ by contradiction. We assume that $\lambda_{\min}(\bar{\bf{G}}^\top \bar{\bf{G}})=1$ without loss of generality. Therefore, we have
\begin{equation}
\begin{aligned}
\rho^2 &\ge ||\F_1*\F_1^\top-\G_1*\G_1^\top||_F^2 + 2||\G_1*\G_2^\top||_F^2 \\
&\ \  + || \F_2*\F_2^\top - \G_2*\G_2^\top  ||_F^2\\
&\ge ||\F_1*\F_1^\top-\G_1*\G_1^\top||_F^2 + || \F_2*\F_2^\top - \G_2*\G_2^\top  ||_F^2 \\
&\ \ + 2\lambda_{\min}(\bar{\bf{G}}_1^\top \bar{\bf{G}}_1)\cdot \lambda_{\max}(\bar{\bf{G}}_2^\top \bar{\bf{G}}_2)/n_3.
\end{aligned}
\end{equation}

Then we prove that if $\sigma_{\min}(\bar{\bf{F}}_1)\ge\sigma_{\max}(\bar{\bf{F}}_2)$, then we have
\begin{equation}
\begin{aligned}
& ||\F_1*\F_1^\top-\G_1*\G_1^\top||_F^2 + || \F_2*\F_2^\top - \G_2*\G_2^\top  ||_F^2 \\
&=\frac{1}{n_3}\left( ||\bar{\bf{F}}_1\bar{\bf{F}}^\top_1-\bar{\bf{G}}_1\bar{\bf{G}}_1^\top||_F^2 + || \bar{\bf{F}}_2\bar{\bf{F}}_2^\top - \bar{\bf{G}}_2\bar{\bf{G}}_2^\top  ||_F^2   \right) \\
&\ \ \ge \underset{d_1,d_2 \in\mathbb{R}_+}{\min} \{ [d_1-\gamma_1]^2 + [d_2-\gamma_2]^2 : d_1\ge d_2 \}.
\end{aligned}
\label{equ:13}
\end{equation}

Consider the following optimization problem
\begin{equation}
\begin{aligned}
 & \underset{\bar{\bf{F}}_1,\bar{\bf{F}}_2}{\min}  \left( ||\bar{\bf{F}}_1\bar{\bf{F}}^\top_1-\bar{\bf{G}}_1\bar{\bf{G}}_1^\top||_F^2 + || \bar{\bf{F}}_2\bar{\bf{F}}_2^\top - \bar{\bf{G}}_2\bar{\bf{G}}_2^\top  ||_F^2   \right):\\
 & \ \ \ \ \lambda_{\min}(\bar{\bf{F}}_1\bar{\bf{F}}_1^\top) \ge  \lambda_{\max}(\bar{\bf{F}}_2\bar{\bf{F}}_2^\top)\} .
\end{aligned}  
\notag
\end{equation}
We relax $\bar{\bf{F}}_1\bar{\bf{F}}_1^\top$ into $\bar{\bf{S}}_1 \succeq 0$ and $\bar{\bf{F}}_2\bar{\bf{F}}_2^\top $ into $\bar{\bf{S}}_2\succeq 0 $ to obtain a low-bound
\begin{equation}
\begin{aligned}
&\ge \underset{\bar{\bf{S}}_1 \succeq 0, \bar{\bf{S}}_2 \succeq 0}{ \min} \{ || \bar{\bf{S}}_1 - \bar{\bf{G}}_1*\bar{\bf{G}}_1^\top ||_F^2 + || \bar{\bf{S}}_2 - \bar{\bf{G}}_2*\bar{\bf{G}}_2^\top ||_F^2:\\
&\ \ \ \ \ \lambda_{\min}(\bar{\bf{S}}_1) \ge \lambda_{\max}(\bar{\bf{S}}_2) \}.
\end{aligned}
\notag
\end{equation}
The problem is invariant to a change of basis, so we change into the eigenbases of $\bar{\bf{G}}_1\bar{\bf{G}}_1^\top$ and $\bar{\bf{G}}_2\bar{\bf{G}}_2^\top$ to obtain 
\begin{equation} 
\begin{aligned}
&= \underset{s_1\ge,s_2\ge }{\min} \{ || s_1 - \lambda\bar{\bf{G}}_1\bar{\bf{G}}_1^\top) ||^2 + || s_2 -\lambda(\bar{\bf{G}}_2\bar{\bf{G}}_2^\top) ||^2:\\ 
&\ \ \ \ \min(s_1) \ge \max(s_2) \},
\end{aligned}
\notag
\end{equation}
where $\lambda(\bar{\bf{G}}_1\bar{\bf{G}}_1^\top),\ \lambda(\bar{\bf{G}}_2\bar{\bf{G}}_2^\top)$ denote the vector of eigenvalues.  We lower-bound this problem by dropping all the terms in the sum of squares except the one associated with $\gamma_1 = \lambda\bar{\bf{G}}_1\bar{\bf{G}}_1^\top)$ and $\gamma_2 = \lambda(\bar{\bf{G}}_2\bar{\bf{G}}_2^\top)$ to yield
\begin{equation}
\ge \underset{d_1,d_2\in\mathbb{R}_+}{\min} \{ [d_1-\gamma_1]^2 + [d_2 - \gamma_2]^2 : d_1\ge d_2  \},
\notag
\end{equation}
which completes the proof of equation (\ref{equ:13}).
By the result of  equation (\ref{equ:13}), if $\gamma_1<\gamma_2$, then $d_1=d_2$ holds at optimality, so the minimum value is $\frac{1}{2}(\gamma_1 - \gamma_2)^2.$ Substituting $\gamma_1 = \lambda_{\min}(\bar{\bf{G}}_1^\top \bar{\bf{G}}_1)$ and $\gamma_2 = \lambda_{\max}(\bar{\bf{G}}_2^\top \bar{\bf{G}}_2)$ then we have
\begin{equation}
\rho^2 \ge \frac{(\gamma_1-\gamma_2)^2}{2n_3} + 2\gamma_1 \gamma_2/n_3 = \frac{1}{2n_3}(\gamma_1 + \gamma_2)^2.
\notag
\end{equation}
But we also have 
\begin{equation}
\begin{aligned}
\gamma_1 + \gamma_2 & = \lambda_{\min}(\bar{\bf{G}}_1^\top \bar{\bf{G}}_1 ) + \lambda_{\max}(\bar{\bf{G}}_2^\top \bar{\bf{G}}_2) \\
&\ge \lambda_{\min}(\bar{\bf{G}}_1^\top \bar{\bf{G}}_1 + \bar{\bf{G}}_2^\top \bar{\bf{G}}_2) = \lambda_{\min}(\bar{\bf{G}}^\top \bar{\bf{G}}) = 1,
\end{aligned} 
\end{equation}
which implies $\rho^2\ge \frac{1}{2n_3}$, a contradiction. Therefore, we complete the proof of Lemma \ref{lemma:9}.
\end{IEEEproof}

Based on the result of Lemma \ref{lemma:9}, we proceed to prove Lemma \ref{lemma:23}.
Firstly we show that the angle between $\G$ and $\F$ satisfies 
\begin{equation}
\begin{aligned}
\sin\beta & \overset{\operatorname{def}}{=} \frac{||(\I-\F*\F^{\dagger})*\G  ||_F} {\sigma_{\min}(\bar{\bf{G}})} 
\overset{(a)}{\le}  \frac{||\G_2||_F}{\sqrt{\lambda_{\min}(\bar{\bf{G}}^\top \bar{\bf{G}})}} \\
&\overset{(b)}{\le}\frac{||\F*\F^\top - \G*\G^\top||_F}{\lambda_{\min}(\bar{\bf{G}}^\top \bar{\bf{G}})} = \rho.
\end{aligned}
\label{equ:15}
\end{equation}
In equation (\ref{equ:15}), inequality (a) uses the fact that 
\begin{equation}
||(\I-\F*\F^{\dagger})*\G||_F \le || (\I-\F_2*\F_2^{\dagger})*\G_2 ||_F \le ||\G_2||_F;
\label{equ:16}
\end{equation}
inequality (b) holds because
\begin{equation}
\begin{aligned}
&|| \F*\F^\top -\G*\G^\top ||_F^2 = ||\F_1*\F_1^\top-\G_1*\G_1^\top||_F^2 \\
&\ \ + || \F_2*\F_2^\top - \G_2*\G_2^\top  ||_F^2  + 2\langle \G_1^\top * \G_1, \G_2^\top * \G_2 \rangle \\
&\ge 2\langle \G_1^\top * \G_1, \G_2^\top * \G_2 \rangle \ge  2\lambda_{\min}(\bar{\bf{G}}_1^\top \bar{\bf{G}}_1)||\G_2||_F^2\\
&\overset{(c)}{\ge}\lambda_{\min}(\bar{\bf{G}}^\top\bar{\bf{G}})||\G_2||_F^2,
\end{aligned}
\end{equation}
where $(c)$ uses  $2\lambda_{\min}(\bar{\bf{G}}_1^\top \bar{\bf{G}}_1) \ge \lambda_{\min}(\bar{\bf{G}}^\top\bar{\bf{G}})$ because 
\begin{equation}
\begin{aligned}
\lambda_{\min}(\bar{\bf{G}}^\top\bar{\bf{G}}) & = \lambda_{\min}(\bar{\bf{G}}_1^\top \bar{\bf{G}}_1 + \bar{\bf{G}}_2^\top\bar{\bf{G}}_2)\le \lambda_{\min} (\bar{\bf{G}}_1^\top\bar{\bf{G}}_1) \\
&+ \lambda_{\max}(\bar{\bf{G}}_2^\top \bar{\bf{G}}_2) \overset{(d)}{\le} 2\lambda_{\min}(\bar{\bf{G}}_1^\top \bar{\bf{G}}_1),
\end{aligned}
\end{equation}
where (d) uses the result of Lemma \ref{lemma:9}.
Then we have 
\begin{equation}
\begin{aligned}
    \sin^2\theta &  = \frac{|| (\I-\F*\F^{-1}) * \bm{\mathcal{M}}_\star *  (\I - \F * \F^{-1}) ||_F^2}{|| \F*\F^\top - \bm{\mathcal{M}}_\star ||_F^2} \\
    & \overset{(1)}{\le} \frac{|| \G_2*\G_2^\top ||_F^2}{|| \F*\F^\top - \bm{\mathcal{M}}_\star ||_F^2} \overset{(2)}{\le} \frac{|| 
\G_2 ||_F^4}{2\langle \G_1^\top *\G_1, \G_2^\top *\G_2 \rangle}\\
&= \frac{||\G_2 ||_F^4}{ 2\langle \bar{\bf{G}}_1*\bar{\bf{G}}_1^\top, \bar{\bf{G}}_2*\bar{\bf{G}}_2^\top \rangle/n_3 } \\
& \overset{(3)}{\le} \frac{|| \G_2 ||_F^4}{2\lambda_{\min}(\bar{\bf{G}}_1^\top \bar{\bf{G}}_1)||\G_2||_F^2} \overset{(4)}{\le} \frac{||\G_2||_F^2}{2[\lambda_{\min}(\bar{\bf{G}}^\top \bar{\bf{G}})-||\G_2 ||_F^2]}\\
& \overset{(5)}{\le} \frac{1}{2} \frac{\rho^2}{1-\rho^2},
\end{aligned}
\end{equation}
where $(1)$ uses the result of equation (\ref{equ:16}); (2) uses the fact that $|| \F*\F^\top - \bm{\mathcal{M}}_\star ||_F \ge 2\langle \G_1^\top *\G_1, \G_2^\top *\G_2 \rangle$; (3) uses $\langle \bar{\bf{G}}_1*\bar{\bf{G}}_1^\top, \bar{\bf{G}}_2*\bar{\bf{G}}_2^\top \rangle/n_3 \ge \lambda_{\min}(\bar{\bf{G}}_1^\top * \bar{\bf{G}}_1)||\G_2||_F^2$; (4) uses $||\G_2||_F^2 \ge \lambda_{\max}(\bar{\bf{G}}_2^\top \bar{\bf{G}}_2)$; (5) uses the assumption (\ref{equ:15}). Therefore, we complete the proof of Lemma \ref{lemma:23}.
\end{IEEEproof}

\begin{lemma}
Suppose that the initialization satisfies $||\X_0-\X_\star||_F\le\rho\sigma_{\min}(\bar{\bf{X}}_\star),\rho=\sqrt{\frac{1}{(1+4\kappa L/\mu)n_3}}$, then we have  
\begin{align*}
    \sin\theta_{\L}^t &:= \frac{||(\I-\P^t_{\L})*\X_\star||_F}{||\L_t*\R_t^\top -\X_\star||_F} \le \frac{\mu}{4L}\\
    \sin\theta_{\R}^t &:= \frac{||\X_\star*(\I-\P^t_{\R})||_F}{||\L_t*\R_t^\top -\X_\star||_F} \le \frac{\mu}{4L},
\end{align*}
where \begin{align*}
    &\P_{\L}^t=\L_t*(\L_t^\top * \L_t)^{\dagger}*\L_t^\top,\\
    &\P^t_{\R}=\R_t*(\R_t^\top *\R_t)^{\dagger}*\R_t^\top.
\end{align*}
\end{lemma}
\begin{IEEEproof}
To prove this Lemma, we need to use the result of Lemma \ref{lemma:23}.
Let $\G =\begin{bmatrix}
    \L_\star\\
    \R_\star
\end{bmatrix}\in\mathbb{R}^{(n_1+n_2)\times r_\star \times n_3}$, $\F=\begin{bmatrix}
    \L\\
    \R
\end{bmatrix}\in\mathbb{R}^{(n_1+n_2)\times r \times n_3}$.
We first ensure that under the assumptions of Lemma \ref{lemma:25}, the assumptions of Lemma \ref{lemma:23} also hold.
In Lemma \ref{lemma:25}, we assume that $$||\X-\X_\star||_F\le\rho\sigma_{\min}(\bar{\bf{X}}_\star),\ \rho=\sqrt{\frac{1}{(1+4\kappa L/\mu)n_3}}.  $$ Combining this with $||\F*\F^\top-\G*\G^\top||_F\le 2||\X-\X_\star||_F$ from Equation (\ref{equ:013}) and the fact that $\sigma_{\min}(\bar{\bf{G}}\bar{\bf{G}}^\top)=\sigma_{\min}(\bar{\bf{G}}^\top\bar{\bf{G}})=2\sigma_{\min}(\bar{\bf{X}}_\star)$, we obtain $$||\F*\F^\top-\G*\G^\top||_F\le \rho\sigma_{\min}(\G*\G^\top),\ \rho=\sqrt{\frac{1}{(1+4\kappa L/\mu)n_3}},$$ which satisfies the assumption in Lemma \ref{lemma:23}.

Note that
\begin{equation}
\begin{aligned}
&||(\I-\F*\F^\dagger)*\G||_F^2 \overset{(a)}{=} \min_{\B\in\mathbb{R}^{r\times r_\star \times n_3}} \left\|  \begin{bmatrix}
    \L_\star\\
    \R_\star
\end{bmatrix} - \begin{bmatrix}
    \L\\
    \R
\end{bmatrix} * \B \right\|_F^2  \\  
&= \min_{\B} \left( ||\L_\star-\L*\B||_F^2+ ||\R_\star-\R*\B||_F^2 \right)\\
&\ge \underset{\B}{\min}||\L_\star-\L*\B||_F^2\overset{(b)}{=}||(\I-\Pi_{\L})*\L_\star||_F^2.
\end{aligned}
\end{equation}
In step (a), we construct
\begin{equation}
\G-\F*\B = \underbrace{(\I-\P_{\F})*\G}_{(1)} + \underbrace{\P_{\F}*\G-\F*\B}_{(2)}. 
\notag
\end{equation}
Since terms (1) and (2) are orthogonal, we have
\begin{align*}
||\G-\F*\B||_F^2 &= ||(\I-\P_{\F})*\G ||_F^2 + ||\P_{\F}*\G-\F*\B||_F^2 \\
& \ge ||(\I-\P_{\F})*\G ||_F^2,
\end{align*}
where the equality holds as long as $\P_{\F}*\G=\F*\B$.

In step (b), we have
\begin{equation}
  \L_\star - \L*\B = \underbrace{(\I-\Pi_{\L})*\L_\star}_{(3)} + \underbrace{\Pi_{\L}*\L_\star-\L*\A}_{(4)}.
  \notag
\end{equation}
Since terms (3) and (4) are orthogonal, we have
\begin{equation}
\begin{aligned}
    ||\L_\star -\L*\B||_F^2 &= ||(\I-\Pi_{\L})*\L_\star||_F^2 + ||\Pi_{\L}*\L_\star-\L*\A||_F^2\\
    &\ge ||(\I-\Pi_{\L})*\L_\star||_F^2,
\end{aligned}
\notag  
\end{equation}
where the equality holds as long as $\Pi_{\L}*\L_\star=\L*\A$.

In order to use the result of Lemma \ref{lemma:23}, we need to establish the inequality between $||(\I-\Pi_{\L})*\X_\star||_F$ and $|| (\I-\F*\F^{\dagger}) * \bm{\mathcal{M}}_\star *  (\I - \F * \F^{\dagger}) ||_F.$ Note that
\begin{equation}
\begin{aligned}
&||(\I-\Pi_{\L})*\X||_F\le ||\R_\star||_2 ||(\I-\Pi_{\L})*\L_\star||_F \\
&\le ||\R_\star||_2 ||(\I-\F*\F^\dagger)*\G||_F \\
&\overset{(a)}{\le}||\R_\star||_2 ||\G^\top*(\I-\P_{\F})*\G||_F/\sigma_{\min}(\bar{\bf{\G}})\\
&\le \sqrt{\kappa/2} ||\G^\top*(\I-\P_{\F})*\G||_F\\
&\le \sqrt{\kappa/2} || (\I-\F*\F^{\dagger}) * \bm{\mathcal{M}}_\star *  (\I - \F * \F^{\dagger}) ||_F,
\end{aligned}
\label{equ:68}
\end{equation}
where $(a)$ use the fact that 
$$
||\G^\top*(\I-\P_{\F})*\G||_F\ge  \sigma_{\min}(\bar{\bf{\G}})||(\I-\F*\F^\dagger)*\G||_F. 
$$
Moreover, combining inequalities (\ref{equ:013}) and (\ref{equ:68}), we have
\begin{align*}
\sin\theta_{\L} &=\frac{||(\I-\Pi_{\L})*\X_\star||_F}{||\L*\R^\top-\X_\star||_F}\le \sqrt{2\kappa} \sin\theta\\
& = \sqrt{2\kappa}\frac{|| (\I-\F*\F^{\dagger}) * \bm{\mathcal{M}}_\star *  (\I - \F * \F^{\dagger}) ||_F}{|| \F*\F^\top - \bm{\mathcal{M}}_\star ||_F} \\
&\le \frac{\sqrt{\kappa}\rho}{\sqrt{1-\rho^2}}\overset{(1)}{\le} \frac{\mu}{4L},
\end{align*}
where (1) use the assumption that $\rho=\sqrt{\frac{1}{(1+4\kappa L/\mu)n_3}}.$
The inequality $$\sin\theta_{\R}^t  = \frac{||\X_\star*(\I-\P^t_{\R})||_F}{||\L_t*\R_t^\top -\X_\star||_F}\le \frac{\mu}{4L}.$$ can be proved by an analogous argument.
\end{IEEEproof}

\subsection{Proof of Lemma 6}
\label{sec:5}

We first state a lemma that relates $\cos\vartheta_{\L}^k$ and $\cos\vartheta_{\R}^k$ to $\sigma_{k+1}^2(\bar{\mathbf L}\bar{\mathbf R}^\top)$.

\begin{lemma}
Under the same assumption of Lemma 22, $\cos\vartheta_{\R}^k$ and $\cos\vartheta_{\L}^k$ defined in Lemma \ref{lemma:011} satisfies
\begin{equation}
\begin{aligned}
&\frac{2(s_{r}^m-k)\sigma^2_{k+1}(\bar{\bf{L}}\bar{\bf{R}}^\top)}{||\bar{\bf{L}}\bar{\bf{R}}^\top-\bar{\bf{X}}_\star||_F^2} - \frac{\mu L}{(L+\mu)^2} \ge \left(\frac{L}{L+\mu}\right)^2 - \cos^2\vartheta_{\L}^k\\
&\frac{2(s_{r}^m-k)\sigma^2_{k+1}(\bar{\bf{L}}\bar{\bf{R}}^\top)}{||\bar{\bf{L}}\bar{\bf{R}}^\top-\bar{\bf{X}}_\star||_F^2} - \frac{\mu L}{(L+\mu)^2} \ge \left(\frac{L}{L+\mu}\right)^2 - \cos^2\vartheta_{\R}^k.
\end{aligned}
\end{equation}
\label{lemma:0012}
\end{lemma}
\begin{IEEEproof}
Note that 
\begin{align*}
&||(\bar{\bf{I}}-\bar{\bf{L}}_k\bar{\bf{L}}_k^\dagger)(\bar{\bf{L}}\bar{\bf{R}}^\top-\bar{\bf{X}}_\star)||_F^2\\
\le &2||(\bar{\bf{I}}-\bar{\bf{L}}_k\bar{\bf{L}}_k^\dagger)\bar{\bf{L}}\bar{\bf{R}}^\top||_F^2 + 2||(\bar{\bf{I}}-\bar{\bf{L}}_k\bar{\bf{L}}_k^\dagger)\bar{\bf{X}}_\star||_F^2\\
\le& 2(s_{r}^m-k)\sigma^2_{k+1}(\bar{\bf{L}}\bar{\bf{R}}^\top) + 2||(\bar{\bf{I}}-\bar{\bf{L}}\bar{\bf{L}}^\dagger)\bar{\bf{X}}_\star||_F^2\\
\end{align*}
and therefore 
\begin{equation}
\sin^2\vartheta_{\L}^k\le \underbrace{\frac{2(s_{r}^m-k)\sigma^2_{k+1}(\bar{\bf{L}}\bar{\bf{R}}^\top)}{||\bar{\bf{L}}\bar{\bf{R}}^\top-\bar{\bf{X}}_\star||_F^2}}_{\Phi_1} + \underbrace{\frac{2||(\bar{\bf{I}}-\bar{\bf{L}}\bar{\bf{L}}^\dagger)\bar{\bf{X}}_\star||_F^2}{||\bar{\bf{L}}\bar{\bf{R}}^\top-\bar{\bf{X}}_\star||_F^2}}_{\Phi_2}.
\label{equ:22}
\end{equation}

For term $\Phi_2$, we have the result of Lemma \ref{lemma:25}: $\Phi_2 \le \frac{\mu}{2L}.$
 Then we rewrite the Equation (\ref{equ:22}) as
 \begin{align*}
    &1-\cos^2\theta_{\L}^k \le \Phi_1 + \frac{\mu}{2L}\\
    \Rightarrow & 1- \frac{\mu}{2L}-\cos^2\theta_{\L}^k \le \Phi_1\\
    \overset{(1)}{\Rightarrow} & \left(\frac{L}{L+\mu}\right)^2+ \frac{\mu L}{(L+\mu)^2} -\cos^2\theta_{\L}^k \le \Phi_1,
 \end{align*}
where (1) uses the fact that
\begin{align*}
1-\frac{\mu}{2L}& = \left(\frac{L}{L+\mu}\right)^2 + \frac{\mu}{2L} \frac{(3L^2-\mu^2)}{(L+\mu)^2} \\
&\ge \left(\frac{L}{L+\mu}\right)^2+ \frac{\mu L}{(L+\mu)^2}.
\end{align*}
Then we have
$$
\frac{2(s_{r}^m-k)\sigma^2_{k+1}(\bar{\bf{L}}\bar{\bf{R}}^\top)}{||\bar{\bf{L}}\bar{\bf{R}}^\top-\bar{\bf{X}}_\star||_F^2} - \frac{\mu L}{(L+\mu)^2} \ge \left(\frac{L}{L+\mu}\right)^2 - \cos^2\vartheta_{\L}^k.
$$
Similarly, we can obtain
$$
\frac{2(s_{r}^m-k)\sigma^2_{k+1}(\bar{\bf{L}}\bar{\bf{R}}^\top)}{||\bar{\bf{L}}\bar{\bf{R}}^\top-\bar{\bf{X}}_\star||_F^2} - \frac{\mu L}{(L+\mu)^2} \ge \left(\frac{L}{L+\mu}\right)^2 - \cos^2\vartheta_{\R}^k.
$$
Therefore, we complete the proof of Lemma \ref{lemma:0012}.

\end{IEEEproof}

\begin{IEEEproof}
When $k=s_{r^\star}^m$, we have
\begin{align*}
\sigma_{s_{r^\star}^m}(\bar{\bf{X}}) &= \sigma_{s_{r^\star}^m}(\bar{\bf{X}}_\star+ \bar{\bf{X}} -\bar{\bf{X}}_\star)   \\
& \overset{(a)}{\ge} \sigma_{s_{r^\star}^m}(\bar{\bf{X}}_\star) - \sqrt{n_3}||\X-\X_\star||_F \\
& \overset{(b)}{\ge} \sqrt{n_3}(\sqrt{1+4\kappa L/\mu}-1) ||\X-\X_\star ||_F.
\end{align*}
where $(a)$ uses the fact that 
\begin{align*}
|\sigma_{r}(\bar{\bf{X}}_1)-\sigma_{r}(\bar{\bf{X}}_2)|&=|\sigma_{r}(\bar{\bf{X}}_1)-\sigma_{r}(\bar{\bf{X}}_2)|\\
&\le ||\bar{\bf{X}}_1-\bar{\bf{X}}_2||_2 \le ||\bar{\bf{X}}_1-\bar{\bf{X}}_2||_F\\
&=\sqrt{n_3} ||\X_1-\X_2||_F;
\end{align*}
(b) uses the assumption that $$||\L*\R^\top-\X_\star ||_F\le \rho \sigma_{\min}(\bar{\bf{X}}_\star),\ \rho=\sqrt{\frac{1}{(1+4\kappa L/\mu)n_3}}.$$
Then we have
\begin{equation}
\frac{\sigma_{s_{r^\star}^m}(\bar{\bf{X}})}{||\X-\X_\star ||_F} \ge  \sqrt{n_3}(\sqrt{1+4\kappa L/\mu}-1)\ge \sqrt{5}-1.
\label{equ:038}
\end{equation}
If $\cos\vartheta_{\L}^{s_{r^\star}^m}\ge \frac{L}{L+\mu}$, then substituting (\ref{equ:038}) into Lemma \ref{lemma:011} with $k=s_{r^\star}^m$ yields 
\begin{equation}
\begin{aligned}
 \frac{||\nabla_{\L}f(\L*\R_t^\top)||_{P_R^*}}{||\L*\R^\top-\X_\star||_F} & \ge  \frac{\mu+L}{2} \left(  \frac{L}{L+\mu} -\varepsilon \right) (1+\lambda/\sigma_{k}^2(\bar{\bf{L}}))^{-\frac{1}{2}} \\
&\ge \frac{\mu}{2} (1+\lambda / \sigma_{s_{r^\star}^m}(\bar{\bf{X}}))^{-\frac{1}{2}}\\
&\ge\frac{\mu}{2}  \left(1+ \frac{\lambda}{(\sqrt{5}-1)||\X-\X_\star ||_F } \right)^{-\frac{1}{2}}.
\end{aligned}
\label{equ:039}
\end{equation}
Otherwise, if $\cos\vartheta_{\L}^{s_{r^\star}^m}< \frac{L}{L+\mu}$, we proceed with an induction argument. Beginning at the base case $k=s_{r^\star}^m$, we use the result of Lemma \ref{lemma:0012} and use $\cos\vartheta_{\L}^{k}< \frac{L}{L+\mu}$ to lower bound $\sigma_{k+1}(\bar{\bf{L}}\bar{\bf{R}}^\top)$
by
\begin{equation}
\begin{aligned}
&\frac{2(s_{r}^m-k)\sigma^2_{k+1}(\bar{\bf{L}}\bar{\bf{R}}^\top)}{||\bar{\bf{L}}\bar{\bf{R}}^\top-\bar{\bf{X}}_\star||_F^2} - \frac{\mu L}{(L+\mu)^2} \ge \left(\frac{L}{L+\mu}\right)^2 - \cos^2\vartheta_{\L}^k\\
&\qquad \qquad \qquad \qquad\qquad\qquad\qquad\ \ >0\\
&\Rightarrow  \frac{\sigma_{k+1}(\bar{\bf{L}}\bar{\bf{R}}^\top)}{||\bar{\bf{L}}\bar{\bf{R}}^\top-\bar{\bf{X}}_\star||_F} \ge \frac{1}{\sqrt{2(s_{r}^m-k)}}  \frac{\sqrt{\mu L}}{L+\mu}.
\end{aligned}
\label{equ:040}
\end{equation}
If $\cos\vartheta_{\L}^{k+1}\ge \frac{L}{L+\mu}$, then substituting (\ref{equ:040}) into Lemma \ref{lemma:011} yields 
\begin{equation}
\begin{aligned}
 &\frac{||\nabla_{\L}f(\L*\R_t^\top)||_{P_R^*}}{||\L*\R^\top-\X_\star||_F} \\
 &\qquad \ge  \frac{\mu+L}{2} \left(  \frac{L}{L+\mu} -\varepsilon \right) (1+\lambda/\sigma_{k+1}^2(\bar{\bf{L}}))^{-\frac{1}{2}}\\
 &\qquad \ge \frac{\mu}{2} \left(1+ \lambda\cdot  \frac{ {\sqrt{2(s_r^m-s_{r_\star}^m)}(L+\mu)} }{\sqrt{\mu L}||\bar{\bf{L}}\bar{\bf{R}}^\top-\bar{\bf{X}}_\star||_F}  \right)^{-\frac{1}{2}}\\
 &\qquad=\frac{\mu}{2} \left(1+ \lambda\cdot  \frac{ {\sqrt{2(s_r^m-s_{r_\star}^m)}(L+\mu)} }{\sqrt{\mu Ln_3}||\L*\R^\top-\X_\star||_F}  \right)^{-\frac{1}{2}}.
\end{aligned}
\label{equ:041}
\end{equation}
Otherwise, if $\cos\vartheta_{\L}^{k+1}< \frac{L}{L+\mu}$, we repeat the same argument in (\ref{equ:040}) with $k\gets k+1$, until we arrive at $k=s_{r}^m$. At this point, Lemma \ref{lemma:0012} guarantees $\cos\vartheta_{\L}^{s_{r}^m}\ge \frac{L}{L+\mu}$, since
\begin{equation}
\begin{aligned}
  \left(\frac{L}{L+\mu}\right)^2 - \cos^2\vartheta_{\L}^k &\le  \frac{2(s_{r}^m-k)\sigma_{k+1}(\bar{\bf{L}}\bar{\bf{R}}^\top)}{||\bar{\bf{L}}\bar{\bf{R}}^\top-\bar{\bf{X}}_\star||_F^2} - \frac{\mu L}{(L+\mu)^2}\\
  &\overset{(1)}{\le} 0 - \frac{\mu L}{(L+\mu)^2}\\
  \Rightarrow  \cos^2\vartheta_{\L}^k &\ge  \left(\frac{L}{L+\mu}\right)^2 +  \frac{\mu L}{(L+\mu)^2},
\end{aligned}
\end{equation}
where $(1)$ uses the fact that $k=s_{r}^m.$ Therefore, the induction step terminates with (\ref{equ:041}). Then, combining the two lower bounds (\ref{equ:039}) and (\ref{equ:041}) via $\min\{a^{-1},b^{-1}\}\ge (a+b)^{-1}$, we have
\begin{align*}
\frac{||\nabla_{\L}f(\L*\R_t^\top)||_{P_R^*}}{||\L*\R^\top-\X_\star||_F} \ge \frac{\mu}{2} \left(2+ c_1 \frac{\lambda }{||\L*\R^\top-\X_\star||_F}   \right)^{-\frac{1}{2}}.  
\end{align*}
where $$c_1=\left(\frac{1}{\sqrt{5}-1}+ \frac{\sqrt{2(s_r^m-s_{r_\star}^m)}(L+\mu)}{\sqrt{\mu Ln_3}}\right).$$
Combining Lemma \ref{lemma:1.3}, we have
\begin{align*}
\frac{||\nabla_{\L}f(\L*\R_t^\top)||_{P_R^*}}{[f(\X)-f(\X_\star)]^{\frac{1}{2}}} \ge \sqrt{\frac{\mu^2}{2L}} \left(2+ c_1 \frac{\lambda }{||\L*\R^\top-\X_\star||_F}   \right)^{-\frac{1}{2}}.  
\end{align*}

As for $\frac{||\nabla_{\R}f(\L*\R_t^\top)||_{P_L^*}}{[f(\X)-f(\X_\star)]^{\frac{1}{2}}}$, we have
\begin{align*}
\frac{||\nabla_{\R}f(\L*\R_t^\top)||_{P_L^*}}{[f(\X)-f(\X_\star)]^{\frac{1}{2}}} \ge \sqrt{\frac{\mu^2}{2L}}  \left(2+ c_1 \frac{\lambda }{||\L*\R^\top-\X_\star||_F}   \right)^{-\frac{1}{2}}.  
\end{align*} by an analogous argument.
\end{IEEEproof}

\subsection{Proof of Lemma 9}
\label{sec:6}
\begin{IEEEproof}

Define $\E_0=\X_\star-\X_0$, and note that $\operatorname{rank}_t(\E)\le 2r$. 
Define the partial tensor Frobenius norm as:
$$
||\X||_{F,r} :=\underset{||\H||_F=1,\operatorname{rank}_t(\H)\le r}{\max} \langle \X, \H\rangle.
$$

Then we have
\begin{align*}
&||\E_0||_F = ||\E_0||_{F,2r}=||\X_0-\X_\star||_{F,2r}\\
&\le || \X_0-\M^*(\y) ||_{F,2r} + || \M^*(\y)-\X_\star ||_{F,2r}\\
&\overset{(1)}{\le} 2|| \M^*(\M(\X_\star))-\X_\star||_{F,2r}\\
&\overset{(2)}{\le} 2\delta||\X_\star||_F,
\end{align*}
where (1) uses the fact that $\X_0 = \arg \underset{\X\in{R}^{n_1\times n_2 \times n_3}}{\min}||\X-\M^*(\y)||_F$; (2) uses the T-RIP condition.
Note that $||\X_\star||_F\le \sqrt{r_\star} \kappa \sigma_{s_{r_\star}^m}(\bar{\bf{X}}_\star)$, then choose $\delta\le \frac{1}{\sqrt{8r\kappa^2(1+12\kappa)n_3}}$, we obtain
$
||\E_0||_F\le \rho \sigma_{s_{r_\star}^m}(\bar{\bf{X}}_\star).
$
\end{IEEEproof}

\end{document}